\documentclass[twoside, 11pt]{article}

\usepackage{mathrsfs}
\usepackage{jmlr2e}
\usepackage{amssymb, amsmath, comment, xspace, enumerate}
\usepackage{graphics, graphicx, subfigure}
\usepackage{microtype,color}
\usepackage[normalem]{ulem}
\usepackage[makeroom]{cancel}
\usepackage{parskip}
\usepackage{mathrsfs}
\usepackage{latexsym}
\newcommand{\energy}{\phi}

\def\<{\begin{equation}}
\def\>{\end{equation}}

\usepackage{csquotes}

\DeclareMathOperator*{\expectation}{\mathbb{E}}
\DeclareMathOperator*{\argmin}{\rm{argmin}}

\ShortHeadings{Neural Empirical Bayes}{Saremi and Hyv\"arinen}

\usepackage{lastpage}
\jmlrheading{20}{2019}{1-\pageref{LastPage}}{3/19; Revised
9/19}{11/19}{19-216}{Saeed Saremi and Aapo Hyv\"arinen}

\begin{document}

\title{Neural Empirical Bayes}

\author{\name Saeed Saremi \email saeed@berkeley.edu \\
       \addr Redwood Center for Theoretical Neuroscience\\
       University of California\\
       Berkeley, CA 94720-3198, USA\\
       NNAISENSE Inc., Austin, TX
       \AND
       \name Aapo Hyv\"arinen \email aapo.hyvarinen@helsinki.fi \\
       \addr        University College London, UK\\ Universit\'e Paris-Saclay, Inria, France\\ University of Helsinki, Finland}

\editor{Yoshua Bengio}

\maketitle



\begin{abstract}%
We unify \emph{kernel density estimation} and \emph{empirical Bayes} and address a set of problems in unsupervised machine learning with a geometric interpretation of those methods, rooted in the \emph{concentration of measure} phenomenon. Kernel density is viewed symbolically as $X\rightharpoonup Y$ where the  random variable $X$ is smoothed to $Y= X+N(0,\sigma^2 I_d)$, and empirical Bayes is the machinery to denoise in a \emph{least-squares} sense, which we express as $X \leftharpoondown  Y$. A learning objective is derived by combining these two, symbolically captured by $X \rightleftharpoons Y$. Crucially, instead of using the original nonparametric estimators, we parametrize  \emph{the energy function} with a neural network denoted by $\phi$; at optimality, $\nabla \phi \approx -\nabla \log f$ where $f$ is the density of $Y$. The optimization problem is abstracted as interactions of high-dimensional spheres which emerge due to the concentration of isotropic Gaussians. We introduce two algorithmic frameworks based on this machinery: (i) a ``walk-jump'' sampling scheme that combines Langevin MCMC (walks) and empirical Bayes (jumps), and  (ii) a probabilistic framework for \emph{associative memory}, called NEBULA, defined \`{a} la Hopfield by the \emph{gradient flow} of the learned energy to a set of attractors. We finish the paper by reporting the \emph{emergence} of very rich {``creative memories''} as attractors of NEBULA for highly-overlapping spheres. 

\end{abstract}

\maketitle
\section{Introduction} \label{sec:intro}
Let $X_1,\dots,X_n$  be an \emph{i.i.d.} sequence in $\mathbb{R}^d$ from a probability density function $f_X$. A well-known \emph{nonparametric estimator} of $f_X$ is the \emph{kernel density estimator} $\widehat{f}$ that spreads the \emph{Dirac masses} of the empirical distribution with a (gaussian) kernel $f_N$~\citep{parzen1962estimation}. Consider $\widehat{f}$ and its standard definition, but take the kernel bandwidth to be a \emph{hyperparameter}. Now, we can consider $\widehat{f}$ to be an estimator of the smoothed density $f_Y = f_X * f_N$, which is the probability density function of the noisy random variable $$Y= X+N(0,\sigma^2 I_d).$$ 
This setup is denoted symbolically as $X\rightharpoonup Y$. 

\begin{remark}[$X$ and $Y$] \label{remark:notation}
The distinction between $X$ and $Y$ is crucial, and in this work we constantly go ``back and forth'' between them. Regarding their probability density functions, we adopt a ``clean notation'': $f(y)= f_Y(y)$ and $f(x) = f_X(x)$, where the subscripts in $f_X$ and $f_Y$ are understood from their arguments $x$ and $y$. In the absence of arguments, the subscripts are brought back, e.g. \begin{equation} \nonumber \label{eq:conv}
	f_Y = f_X *f_N.
\end{equation}
In addition, as it will become clear shortly, we are only concerned with approximating $\nabla \log f_Y$ in this paper, therefore the subscript $Y$ becomes the default; it is understood that $f=f_Y$.
\end{remark}

Next, we discuss the random variable $Y$ in $\mathbb{R}^d$ from the perspective of \emph{concentration of measure}. Assume that the random variable $X$ in high dimensions is concentrated on a low-dimensional manifold $\mathscr{M}$, and denote the manifold where the random variable $Y= X+N(0,\sigma^2 I_d)$ is concentrated as $\mathscr{N}$.\footnote{Unfortunately, the concept of ``manifold'' for a probability distribution, let alone its dimension, is quite a fuzzy concept. But it should also be stated that this is a very important problem; the \emph{assumption} on the \emph{existence} of $\mathscr{M}$ is a pillar of machine learning~\citep{saul2003think, bengio2013representation}.} We are interested in formalizing the hypothesis that as $d \rightarrow \infty$, the convolution of $f_X$ and $f_N$ (for any $\sigma$) would ``disintegrate'' $\mathscr{M}$ such that $\text{dim}(\mathscr{N})\gg \text{dim}(\mathscr{M}).$ We study an example of a ``gaussian manifold'', defined by $X \sim N(0,\Sigma_\sharp)$ where $$
  \Sigma_\sharp = \begin{bmatrix} 
          I_{d_\sharp} & 0 \\
          0 & \epsilon^2 I_{d-d_\sharp} \\
        \end{bmatrix}, 
$$ with $\epsilon \ll 1$ and $d_\sharp \ll d$. We revisit textbook calculations on concentration of measure and show that in high dimensions, the smoothed manifold $\mathscr{N}$ is approximately a sphere of dimension $d-1$: $$\mathscr{N} \approx \sqrt{(\epsilon^2+\sigma^2) (d-d_{\sharp})} S^{d-1}.$$ This analysis paints a picture of ``manifold disintegration-expansion'' as a general phenomenon. 


\begin{itemize}
	\item[(C1)] The first conceptual contribution of this paper is to expose a notion of manifold disintegration-expansion, with a general takeaway that in (very) high dimensions, gaussian smoothing pushes away (all) the probability masses near $\mathscr{M}$. This is due to the fact that in high dimensions, the Gaussian random variable $$N(0,\sigma^2 I_d) \approx {\rm Unif}(\sigma \sqrt{d}S^{d-1}) $$ is not necessarily concentrated on $\mathscr{M}$, and there are  many directions (asymptotically infinite) where samples from $Y$ ``escape'' the manifold. The thesis is that, in convolving $f_X$ and $f_N$, $\mathscr{M}$ would be mapped to a (much) higher dimensional manifold.
\end{itemize} 

A seemingly unrelated theory is {\it empirical Bayes}, as formulated by~\citet{robbins1956empirical}, which we use for pulling the probability mass back towards $\mathscr{M}$. In 1956, Robbins considered a scenario of a random variable $Y$ that depends ``in a known way'' on an ``unknown'' random variable $X$.\footnote{\label{foot:xy} The starting point in~\citep{robbins1956empirical} was the existence of a noisy random variable that was denoted by $X$. In that setup, one can \emph{only} observe values of $X$. Here, we started with the \emph{``clean''} i.i.d. sequence $X_1,\dots,X_n$ and \emph{artifically created} $Y=X+N(0,\sigma^2 I_d)$. This departure in starting points is related to our new take on empirical Bayes which will become clear in Section~\ref{sec:deen}.} He found that, given an observation $Y=y$, the \emph{least squares estimator} of $X$ is the \emph{Bayes estimator}, and quite remarkably, can be expressed \emph{purely} in terms of the distribution of $Y$ (he showed this for Poisson, geometric, and Laplacian kernels). Robbins' results were later extended to \emph{gaussian kernels} by~\citep{miyasawa1961empirical} who derived the estimator\begin{equation}\nonumber \widehat{x}(y) = y + \sigma^2 \nabla \log f(y).\end{equation}
The least-squares estimator of $X$ above is derived for  \emph{any} $\sigma$, not necessarily infinitesimal. To signify this important fact, we also refer to this estimation as  ``Robbins jump'' or \emph{jump} for short. The empirical Bayes machinery is denoted symbolically as $$X \leftharpoondown  Y.$$ 


 The smoothing of kernel density estimation, $X\rightharpoonup Y$, and the denoising mechanism of empirical Bayes, $X \leftharpoondown  Y$, come together to define the learning objective $$ 	\mathcal{L}=\expectation \Vert X - \widehat{x}(Y)\Vert^2. $$
(The squared $\ell_2$ norm $\Vert \cdot \Vert^2$ in the definition of $\mathcal{L}$ is due to the fact that the estimator of $X$ is a {\it least-squares estimator}.) In this setup, we take the two \emph{nonparametric estimators} that defined $\mathcal{L}$ and parametrize the \emph{energy function} of the random variable $Y$ with a neural network $\energy: \mathbb{R}^d \rightarrow \mathbb{R}$ with parameters $\theta$. It should be stated that $\energy$ is defined modulo an additive constant. The learning objective, expressed in terms of $\theta$, is therefore given by
\begin{equation}\label{eq:deen}
	\mathcal{L}(\theta) = \expectation \Vert X - Y+\sigma^2 \nabla \energy(Y,\theta)\Vert^2.
\end{equation} 
{\it (Throughout the paper, $\nabla$ is the gradient taken with respect to the inputs $y$, not parameters.)}


\begin{itemize}
	\item[(C2)]The second conceptual contribution of this paper is this unification of kernel density estimation  and empirical Bayes. The unification, denoted symbolically as $X \rightleftharpoons Y$, is encapsulated in the expression for the learning objective above, $\mathcal{L}=\expectation \Vert X - \widehat{x}(Y)\Vert^2.$ We thus combine two principles of nonparametric estimation into a single learning objective. In optimizing the objective, we choose to parametrize the energy function with a (overparametrized) neural network. The growing understanding on the representational power of deep (and wide) neural networks and  the effectiveness of SGD for the ``problem of learning''~\citep{vapnik2013nature} is behind this choice.
\end{itemize}

\begin{remark}[DEEN]
For gaussian noise, the objective $\mathcal{L}=\expectation \Vert X - \widehat{x}(Y)\Vert^2$ is the same as the learning objective in ``deep energy estimator networks'' (DEEN)~\citep{saremi2018deep} which itself was based on  denoising score matching~\citep{hyvarinen2005estimation, vincent2011connection}. However, this equivalence breaks down beyond gaussian kernels\textemdash see~\citep{raphan2011least} for a comprehensive survey of empirical Bayes least squares estimators. From this angle, empirical Bayes appears as a more fundamental framework to formulate the problem of unnormalized density estimation for noisy random variables. 
\end{remark}

\begin{figure}[h!]
\begin{center}
    \begin{subfigure}[$d=2$ gaussian]{\includegraphics[width=0.26\textwidth]{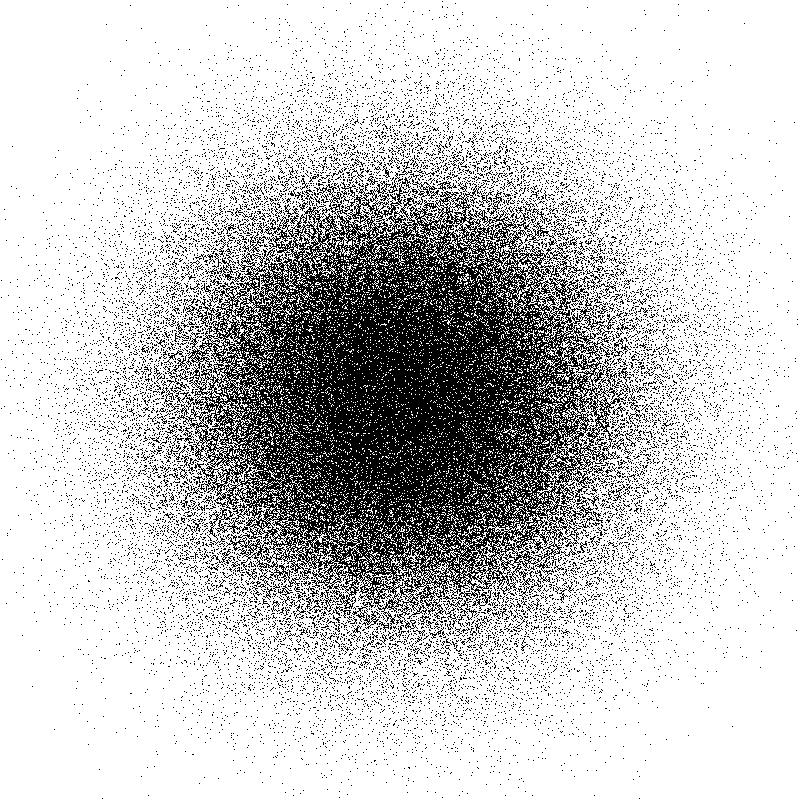}}
    \end{subfigure}
    \begin{subfigure}[$d\gg1$ gaussian]{\includegraphics[width=0.26\textwidth]{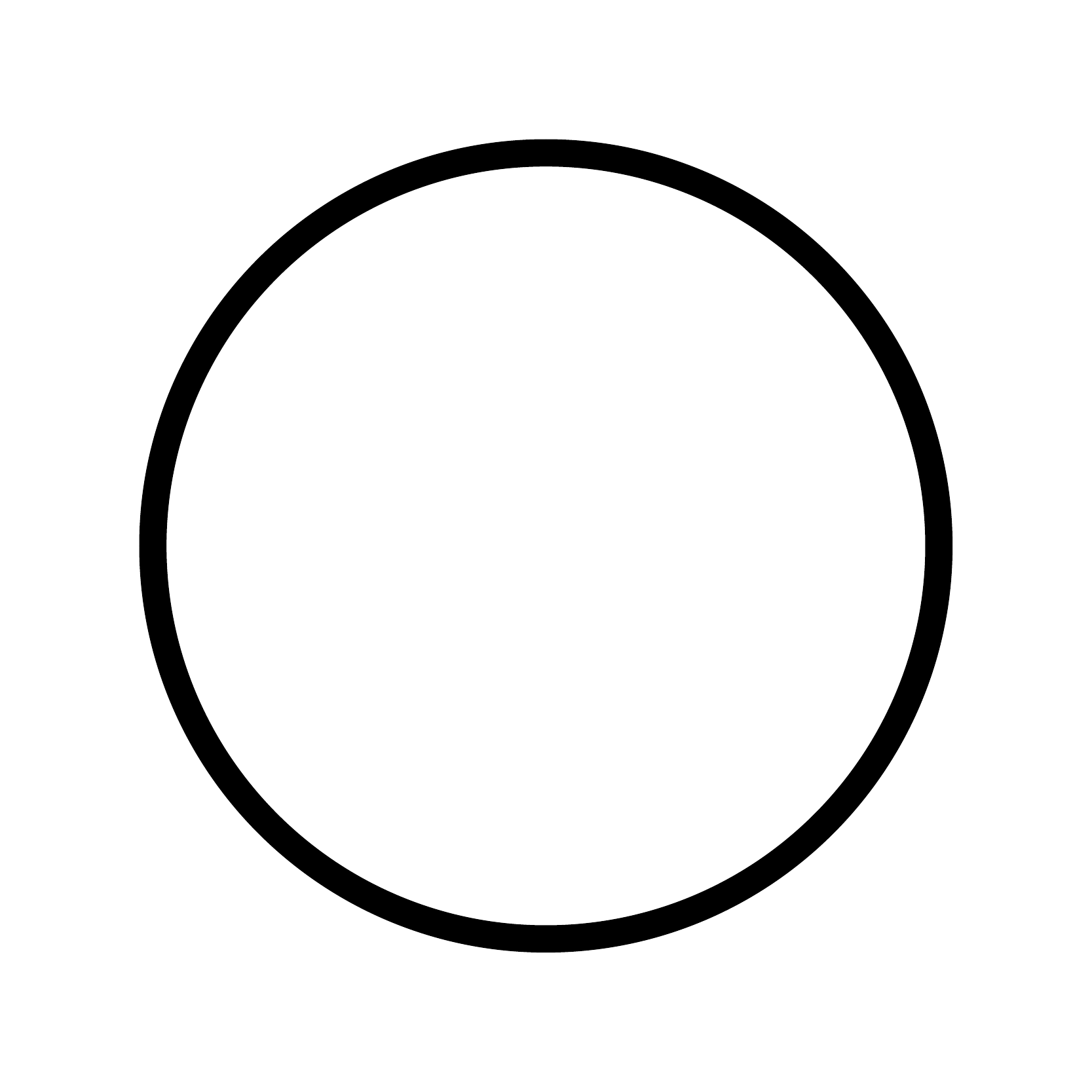}}%
    \end{subfigure}
    \begin{subfigure}[$X \rightleftharpoons Y$ ]{\includegraphics[width=0.26\textwidth]{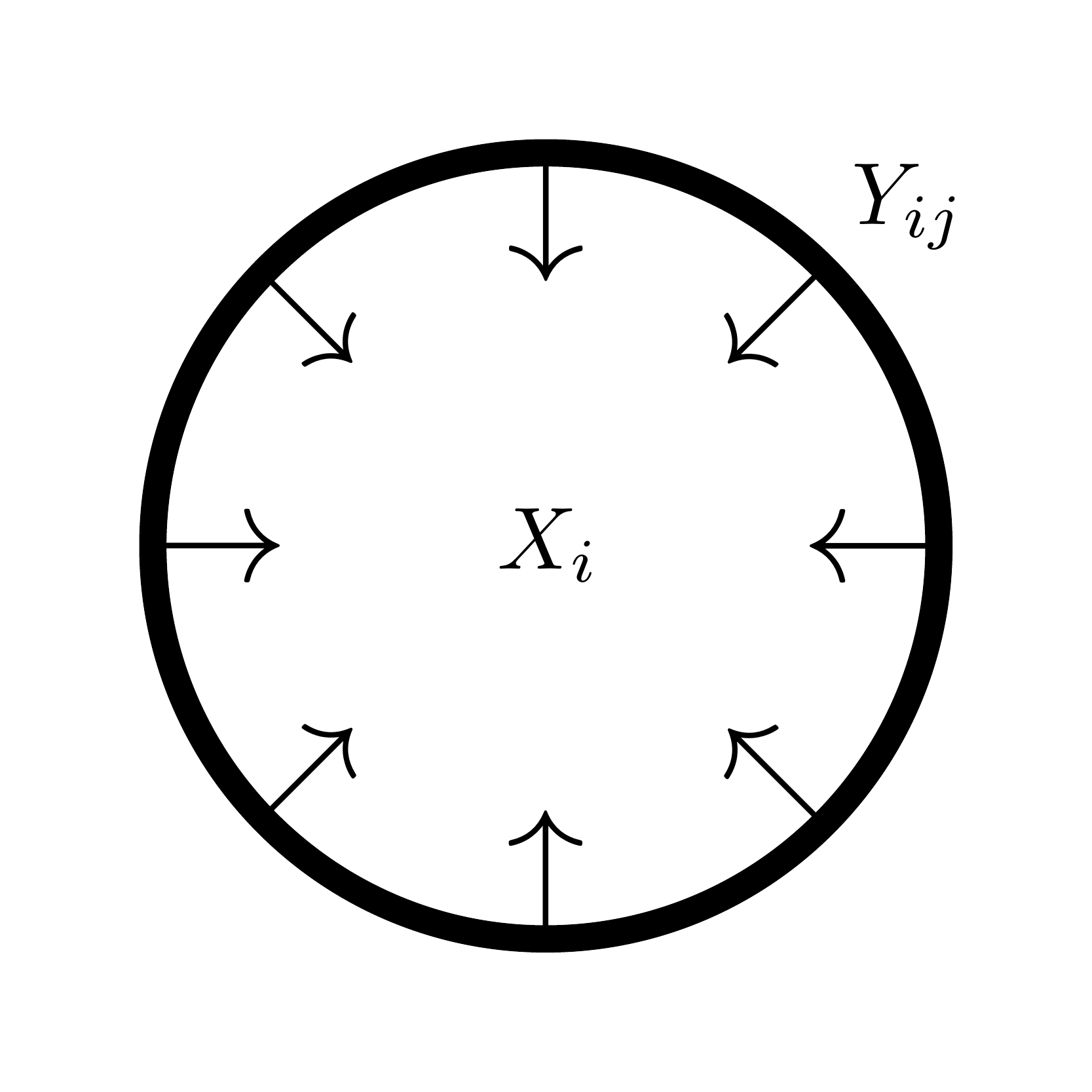}}
    \end{subfigure}    
    \caption{(a) Samples from a 2D isotropic gaussian, obtained and rendered in the programming language Processing. (b) Schematic of an isotropic gaussian in high dimensions, where the concentration of norm is illustarted. (c) Schematic of the $i$-sphere, with samples $Y_{ij}=X_i+\varepsilon_j,~\varepsilon_j \sim N(0,\sigma^2 I_d)$. The arrows represent $-\nabla \energy$, evaluated on the sphere. The learning objective is  encapsulated by $X \rightleftharpoons Y$, where the squared $\ell_2$ norm $\Vert X_i - \widehat{x}(Y_{ij})\Vert^2$ is the learning signal and minimized in expectation. Ignoring the other spheres,  the learning objective is constructed such that $-\nabla \phi$ evaluated at $Y_{ij}$ points to $X_i$.}
    \label{fig:gaussians}
\end{center}
\end{figure}

When analyzing a finite i.i.d. sequence, $X_1,\dots,X_n$, i.i.d. samples from $Y= X+N(0,\sigma^2 I_d)$ are generated as
$$ Y_{ij} = X_i + \varepsilon_j,\ \varepsilon_j \sim N(0,\sigma^2 I_d),$$
and the learning objective $\mathcal{L}(\theta)$ is then approximated as
$$
        \mathcal{L}(\theta) \approx \sum \left\Vert X_i - Y_{ij}+\sigma^2 \nabla \energy(Y_{ij},\theta)\right\Vert^2,
$$
where $\sum$ is a shorthand for $\frac{1}{nm}\sum_{i=1}^n \sum_{j=1}^m$.  In high dimensions, the samples $Y_{ij}$ are (approximately) uniformly distributed on a ``thin spherical shell'' around the sphere $\sigma \sqrt{d} S_i^{d-1}$ with its center at $X_i$ which leads to the following definition. 
\begin{definition}[$i$-sphere] \label{def:isphere}
The sphere $\sigma \sqrt{d} S_i^{d-1}$ centered at $X_i$ is a useful abstraction and it is referred to as ``$i$-sphere''. The $i$ in $i$-sphere refers to the index $i$ in $X_i$, and the index $j$ is reserved for the gaussian noise $\varepsilon_j$. Therefore, $Y_{ij}$ is the $j$th sample on the $i$-sphere. Fixing the index $i$, the samples $Y_{ij}$ are approximately uniformly distributed on a thin spherical shell around the $i$-sphere (see Figure~\ref{fig:gaussians}).
\end{definition}
 In fact, the learning based on $\mathcal{L}$ can be interpreted as shaping the global energy function such that locally, for samples  $Y_{ij}$ from a single $i$-sphere, the denoised versions $\widehat{x}(Y_{ij})$ come as close as possible (in squared $\ell_2$ norm) to $X_i$ at the center (see Figure~\ref{fig:gaussians}c). This ``shaping of the energy function'' is programmed in a simple algorithm,\begin{quote}
\centering
	{\it DEEN: minimize $\mathcal{L}(\theta)$ with stochastic gradient descent and return $\theta^*$.}
\end{quote}



\begin{remark}
	 In most of the paper, we are interested in studying $\phi$ which is already at optimality  with parameteres $\theta^* = \argmin \mathcal{L}(\theta)$. In these contexts, $\theta^*$ is dropped, and its presence is understood, e.g. at optimality, $\phi \approx - \log f$ (modulo a constant). 
\end{remark}

Next, we define a notion of interactions between $i$-spheres that we find useful in thinking about the goal of approximating the \emph{score function} through optimizing $\mathcal{L}(\theta)$. However, this can be skipped until our discussions of the associative memory in Sections~\ref{sec:NEBULA} and~\ref{sec:creative}.
 
\begin{definition}[$i$-sphere interactions] \label{def:isi}
	The interaction between $i$-spheres is defined as the competition that emerge due to the presence of $\nabla \phi$  in the problem of minimizing $\mathcal{L}$. In other words, it refers to ``the set of constraints'' that $\nabla \phi$, the gradient of the globally defined energy function, has to satisfy locally on $i$-spheres to arrive at the optimality such that $\nabla \phi \approx -\nabla \log f.$ It is indeed an abstract notion in this work, but it is a useful abstraction to have which we come back to in thinking about the gradient flow $y'=-\nabla \phi(y).$ It should be stated that there is also a notion of interaction that already exists for the kernel density in the sense of ``mass aggregation'' (see Section~\ref{sec:creative} for references), but in this framework, $i$-spheres interact even when they do not overlap in that they ``communicate'' via $\nabla \energy$. 
	\end{definition}
	
\begin{itemize}
	\item[(C3)]The third conceptual contribution of this paper is introducing the physical picture of interactions between \emph{i.i.d.} samples. This is a useful abstraction in this work in thinking about the problem of approximating the \emph{score function} with the gradient of an \emph{energy function}. The interaction is a code for ``the set of constraints'' that $\nabla \phi$ (evaluated at i.i.d. samples $Y_{ij}$) has to satisfy to arrive at optimality, $\nabla \phi \approx -\nabla \log f$. These interactions could also give rise to some \emph{collective phenomena}~\citep{anderson1972more}.
	\end{itemize}

\begin{figure}[h!]
\begin{center}
    \begin{subfigure}[$\sigma < \chi_{ii'}$]{\includegraphics[width=0.23\textwidth]{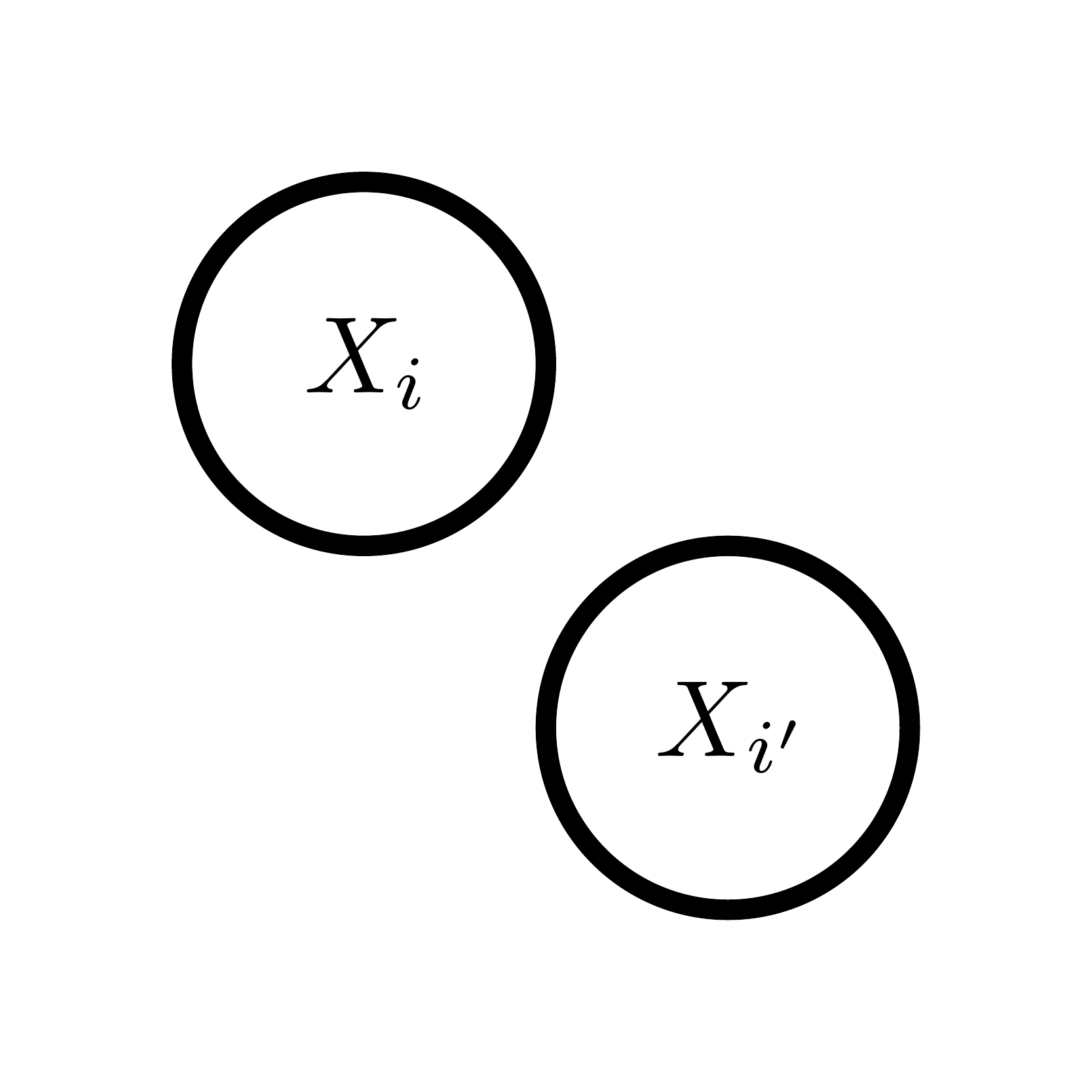}}%
    \end{subfigure}
    \hspace{0.8cm}
    \begin{subfigure}[$\sigma \approx \chi_{ii'}$]{\includegraphics[width=0.23\textwidth]{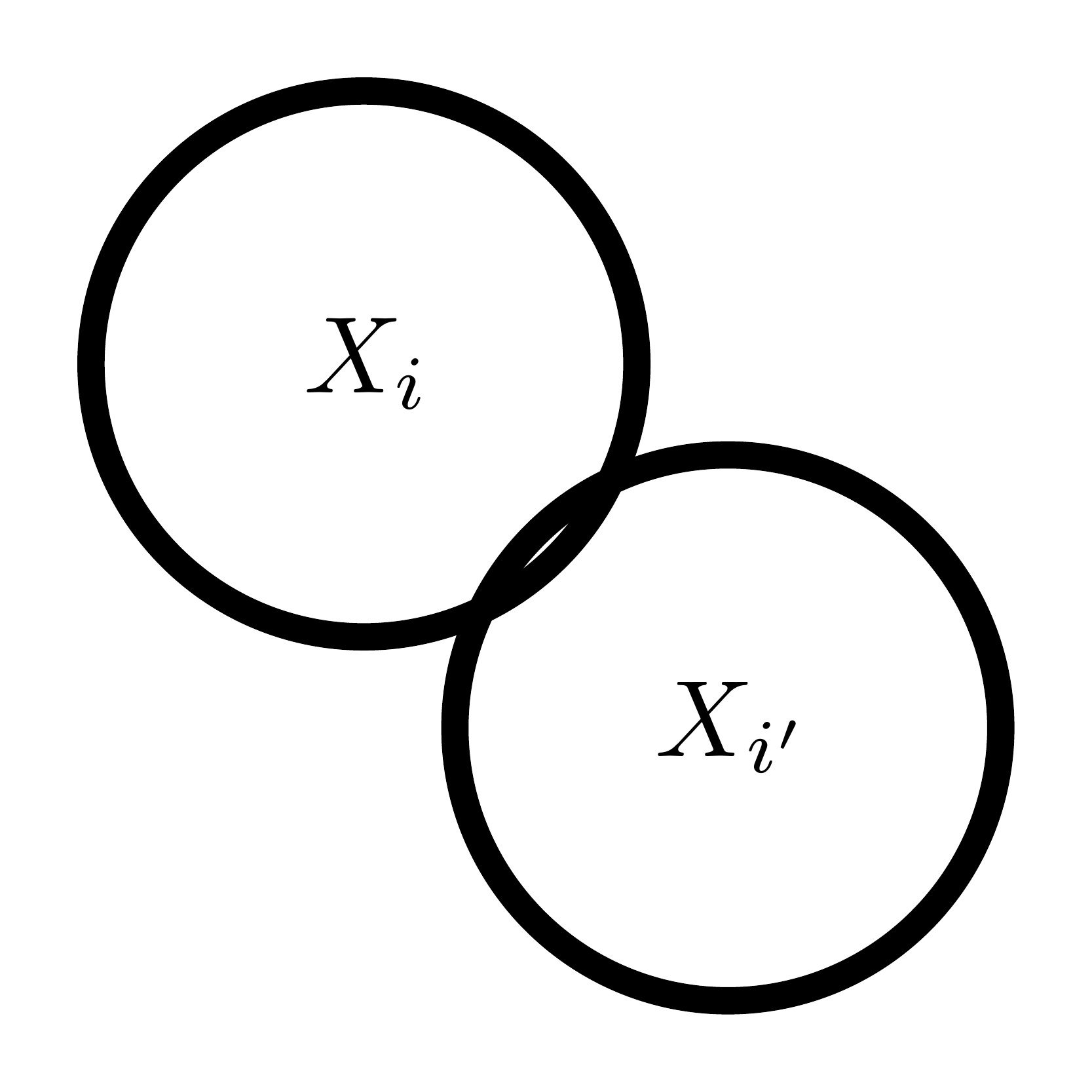}}%
    \end{subfigure}
    \hspace{1cm}
    \begin{subfigure}[$\sigma > 2 \chi_{ii'}$]{\includegraphics[width=0.23\textwidth]{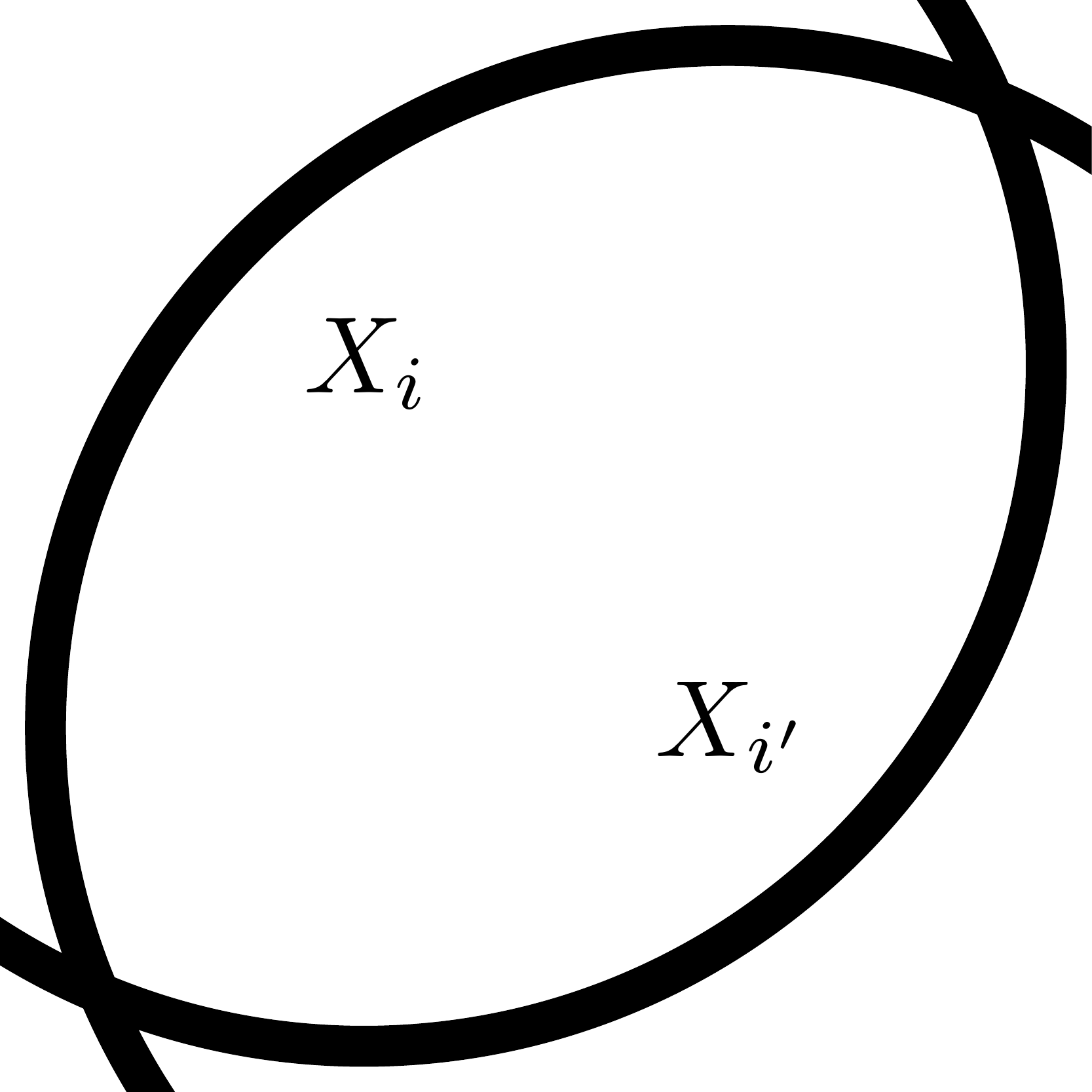}}%
    \end{subfigure}   
    \caption{({\it overlapping $i$-spheres}) The extent of the overlap between $i$-sphere and $i'$-sphere is tuned by $\sigma$ in relation to $\chi_{ii'}=\Vert X_i-X_{i'}\Vert/(2\sqrt{d}).$ The scaling ($2\sqrt{d}$) is due to the fact that $N(0,\sigma^2 I_d) \approx \text{Unif}(\sigma \sqrt{d} S^{d-1})$ in high dimensions.}   
    \label{fig:twospheres}    
\end{center}
\end{figure}

Next, we outline the technical contributions of this paper. Approximating $\nabla \log f$ in the empirical Bayes setup leads to a novel sampling algorithm, a new notion of associative memory and the emergence of ``creative memories''. First, we define the matrix $\chi$ that was introduced to quantify $i$-sphere overlaps and was used for designing experiments.
\begin{itemize}
	\item[(i)] Concentration of measure leads to a geometric picture for the kernel density as a \emph{``mixture of $i$-spheres''}, and we define the matrix $\chi$ to quantify the extent to which  the spheres in the mixture overlap; the entries in $\chi$ are essentially pairwise distances, scaled by $2\sqrt{d}$ (where the scaling is related to the concentration of isotropic Gaussians in high dimensions):\begin{equation} \label{eq:chi} \chi_{ii'} = \frac{\Vert X_i - X_{i'} \Vert}{2\sqrt{d}}.\end{equation} 
	The $i$-sphere and the $i'$-sphere do not overlap if $\sigma<\chi_{ii'}$, and they overlap if $\sigma>\chi_{ii'}$ (see Figure~\ref{fig:twospheres}). Of special interest is $\sigma_c$, defined as $$\sigma_c = \max_{ii'} \chi_{ii'}.$$ We define ``extreme noise'' as the regime $\sigma>\sigma_c$; it is the regime where all $i$-spheres have some degree of overlap.
	
	\item[(ii)]  ``Walk-jump sampling'' is an approximative sampling algorithm that first draws exact samples from the density $\exp(-\phi)/Z$ ($Z$ is the unknown normalizing constant) by \emph{Langevin MCMC}: 
	\begin{equation}\nonumber
	y_{t+1} = y_t - \delta^2 \nabla \energy(y_t) + \sqrt{2}  \delta \varepsilon ,\ ~\varepsilon\sim  N(0,I_d),
\end{equation}
where $\delta \ll 1$ is the step size and here $t$ is discrete time. At an arbitrary time $\tau$, approximative samples ``close'' to $\mathscr{M}$ are generated with the least-squares estimator of $X$\textemdash \emph{the jumps}:
\begin{equation}\nonumber
	\widehat{x}(y_\tau)  = y_\tau - \sigma^2 \nabla \energy(y_\tau).
\end{equation} A signature of walk-jump sampling is that the walks and the jumps are decoupled in that the jumps can be made at arbitrary times.

\item[(iii)]  ``Neural empirical Bayes associative memory''\textemdash named NEBULA, where the ``LA'' refers to {\it\`{a} la}~\citet{hopfield1982neural}\textemdash is defined as the flow to strict local minima of the energy function $\energy$. In continuous time, the memory dynamics is governed by the \emph{gradient flow}:$$ y'(t)=-\nabla \energy(y(t)).$$In retrieving a ``memory'', one flows deterministically to an \emph{attractor}. NEBULA is an intriguing  construct as the deterministic dynamics is governed by a \emph{probability density function}, since $ \nabla \phi \approx -\nabla \log f$\textemdash this is very far from true for Hopfield networks.

\item[(iv)] Memories in NEBULA are believed to be formed due to {\it $i$-sphere interactions} (Definition~\ref{def:isphere}). We provide evidence for the presence of this abstract notion of interaction in this problem by reporting the emergence of very structured memories in a regime of highly overlapping $i$-spheres. They are named ``creative memories'' because they have intuitively appealing structures, while clearly being new instances, in fact quite different from the training data.
\end{itemize}

This paper is organized as follows. \emph{Manifold disintegration-expansion} is discussed in Section~\ref{sec:hdp}. In Section~\ref{sec:deen}, we review empirical Bayes and give a derivation of the least squares estimator of $X$ for gaussian kernels. We then talk about a \emph{Gedankenexperiment}\textemdash in the school of~\citet{robbins1956empirical} and \citet{robbins1951stochastic}\textemdash to bring home the \emph{unification scheme} $X\rightleftharpoons Y$ and shed some light on the inner-working of the learning objective. In Section~\ref{sec:deendenoise}, we define the notion of \emph{extreme noise} and demonstrate two extreme denoising experiments. In Section~\ref{sec:walkjump}, we present the \emph{walk-jump sampling} algorithm. \emph{NEBULA} is defined in Section~\ref{sec:NEBULA}, where the abstract notion of \emph{$i$-sphere interactions} is grounded to some extent; we present two sets of experiments with qualitatively different behaviors. In Section~\ref{sec:creative}, we report the emergence of \emph{creative memories} for two values of $\sigma$ close to $\sigma_c$. We finish with a summary.

 \vspace{0.3cm}
\setcounter{subfigure}{0}
\section{Manifold disintegration-expansion in high dimensions} \label{sec:hdp}Our low-dimensional intuitions break down spectacularly in high dimensions. This is in large part due to the exponential growth of space in high dimensions (this abundance of space also underlies the {\it curse of dimensionality}, well known in statistics and machine learning). Our focus in this section is to develop some intuitions on the effect of gaussian smoothing, $X\rightharpoonup Y$, on the data manifold $\mathscr{M}$. We start by a summary of results on {\it concentration of measure}. The textbooks~\citep{ledoux2001concentration, tao2012topics,vershynin2018high} should be consulted to fill in details. We then extend the textbook calculations for a ``gaussian manifold'' and make a case for \emph{manifold disintegration-expansion}.

Start with the isotropic gaussian. In 2D, samples from the gaussian form a ``cloud of points'', centered around its mode as illustrated in Figure~\ref{fig:gaussians}a. Next, take the isotropic Gaussian in $d$ dimensions, $(X_1, \dots, X_d) = N(0,I_d)$, and ask where the random variable is likely to be located. Before stating a well-known result on the concentration of measure, we can build intuition by observing the following identities for the expectation and the variance of the squared norm:
\begin{eqnarray*}
	\expectation \Vert X \Vert^2 &=& \sum_{i=1}^d \expectation  X_i^2 = d,\\
    \mathbb{V} \Vert X \Vert^2  &=& \sum_{i=1}^d \expectation(X_i^2-1)^2 = 2d.	
\end{eqnarray*}
Since the components of $X=(X_1,\dots,X_d)$ are jointly independent, and since $d \gg 1$, the following holds with high probability,
\begin{eqnarray*}
	\Vert X \Vert^2 &=& d\pm O(\sqrt{d}),\\
	\Vert X \Vert_{\ } &=& \sqrt{d} \pm O(1).
\end{eqnarray*}
 The concentration of norm is visualized in Figure~\ref{fig:gaussians}b; also see Figure 3.2 in~\citep{vershynin2018high}. In contradiction to our low-dimensional intuition, in high dimensions, the Gaussian random variable $N(0,I_d)$ is not concentrated close to the mode of its density $f_N$ (at the origin here), but in a ``thin spherical shell'' of width $O(1)$ around the sphere of radius $\sqrt{d}$. The concentration of norm suggests an even deeper phenomenon, the concentration of measure, captured by a non-asymptotic result for the deviation of the norm $\Vert X\Vert$ from $\sqrt{d}$, where the deviation has a \emph{sub-gaussian} tail~\citep{tao2012topics,vershynin2018high}:\begin{equation}
    \nonumber
	\mathbb{P}\left(\left\vert \Vert X\Vert -\sqrt{d}\right\vert\geq t\right) \leq 2 \exp(-ct^2), \hspace{0.1in} {\rm for\ all\ }t\geq 0.
\end{equation}
A related result states that
\begin{equation}
\nonumber
	\mathbb{P}(\Vert X \Vert \leq \epsilon \sqrt{d}) \leq (C\epsilon)^d,\hspace{0.1in} {\rm for\ all\ }\epsilon\geq 0.
\end{equation}
In above expressions, $C$ and $c$  are absolute constants. In high dimensions, the Gaussian random variable is thus approximated with the uniform distribution on the sphere $\sigma\sqrt{d} S^{d-1}$,
\begin{equation}\nonumber
	N(0,\sigma^2 I_d) \approx {\rm Unif}(\sigma\sqrt{d} S^{d-1}).
\end{equation}
The approximation becomes equality in distribution, as $d\rightarrow \infty$. 


\begin{remark}\label{remark:uniform}
In high dimensions, our intuitions for the uniform distribution itself breaks down, where the probability mass no longer has a ``uniform geometry''. For example, ${\rm Unif}([0,1]^d)$ is concentrated near the hyperplane~$ X_1+\dots+X_d = d/2$, which is easy to see, starting from $\expectation X = 1/2$ for the univariate $X$~\citep{ledoux2001concentration}. The same goes for ${\rm Unif}(\sqrt{d} S^{d-1})$, which is a prime example of ``concentration without independence'', with counterintuitive results like the ``blow-up'' phenomenon~\citep{vershynin2018high}.
 \end{remark}



We finish this section with a new analysis. Consider $X \sim N(0,\Sigma_\sharp)$ in $\mathbb{R}^d$. A simple low-dimensional  ``manifold structure'' with the dimension $d_\sharp \ll d$  is imposed by considering
$$
  \Sigma_\sharp = \begin{bmatrix} 
          I_{d_\sharp} &   0 \\
          0 & \epsilon^2 I_{d-d_\sharp} \\
        \end{bmatrix}, 
$$
where $\epsilon \ll 1$. The manifold $\mathscr{M}$ is viewed as a ``gaussian manifold'' where, in an abuse of notation, $\text{dim}(\mathscr{M})\approx d_\sharp$. Now consider $Y= X + N(0,\sigma^2 I_d)$ which means
\begin{equation}
\nonumber
	Y = N(0,\Sigma_\sharp +\sigma^2 I_d).
\end{equation}
We repeat the concentration of norm calculations:
\begin{eqnarray*}
	\expectation \Vert Y \Vert^2 &=& d_{\sharp}(1+\sigma^2)+(d-d_{\sharp})(\epsilon^2+\sigma^2), \\
	\mathbb{V} \Vert Y \Vert^2 &=&  2d_{\sharp}(1+\sigma^2)^2 +2(d-d_{\sharp})  (\epsilon^2+\sigma^2)^2.
\end{eqnarray*}
In high dimensions, assuming
\begin{equation}
\nonumber
(d/d_\sharp-1) \gg \max(\Delta,\Delta^2), \hspace{0.2cm} \Delta= (1+\sigma^2)/(\epsilon^2+\sigma^2),
\end{equation}	
the ``manifold terms'', $d_{\sharp}(1+\sigma^2)$ and $d_{\sharp}(1+\sigma^2)^2$, become negligible, and the following holds with high probability:\begin{eqnarray*}
  \Vert Y \Vert^2 &=& (\epsilon^2+\sigma^2) ( d-d_{\sharp} \pm O(\sqrt{d-d_{\sharp}}) ),  \\
	\Vert Y \Vert_{\ } &=& \sqrt{\epsilon^2+\sigma^2}(\sqrt{d-d_{\sharp}}\pm O(1)).
\end{eqnarray*}
The calculations in the previous page are reproduced by setting $\sigma=1$, $\epsilon=0$, $d_\sharp=0$.

 
In summary, under the convolution $f_Y=f_X*f_N$, the ``gaussian manifold'' $\mathscr{M}$ is transformed/mapped to $$\mathscr{N} \approx \sqrt{(\epsilon^2+\sigma^2) (d-d_{\sharp})} S^{d-1},$$therefore $\text{dim}(\mathscr{N}) \approx d-1$. In our terminology, $\mathscr{M}$ has been \emph{disintegrated-expanded} to $\mathscr{N}$. We expect this phenomenon to hold for any $\mathscr{M}$ asymptotically, i.e. as  $d\rightarrow \infty$, $\mathscr{N} = \sigma \sqrt{d} S^{d-1}$, but (unfortunately) the limit $d\rightarrow \infty$ is not practical. However, the picture should stay: in high dimensions, the convolution of $f_X$ and $f_N$  has severe side effects on the manifold $\mathscr{M}$ itself. Smoothing is indeed desired, but also a mechanism to restore $\mathscr{M}$. In the next section, we discuss such a mechanism using \emph{empirical Bayes}, where the ``restoration'' is in \emph{least squares}.




\section{Neural empirical Bayes}
\label{sec:deen}
In a seminal work from 1956, titled \emph{an empirical Bayes approach to statistics}, Robbins considered an intriguing scenario of a random variable $Y$ having a probability distribution that depends ``in a known way'' on an ``unknown'' random variable $X$~\citep{robbins1956empirical}. Observing $Y=y$, the \emph{least-squares estimator} of $X$ is the \emph{Bayes estimator} (see Remark~\ref{remark:notation}):
\begin{equation}
\nonumber
\widehat{x}(y) = \frac{\int x f(y|x) f(x) dx }{\int f(y|x) f(x) dx} = \int x f(x|y) dx = \expectation(X|Y=y).
\end{equation}
If $f_X$ is known to ``the experimenter'', $\widehat{x}$ is a computable function, but what if the prior $f_X$ is unknown? It is quite remarkable that for a large class of kernels, the least-squares estimator can be derived in closed form purely in terms of the distribution of $Y$. Informally speaking, there is an ``abstraction barrier''~\citep{abelson1985structure} between $X$ and $Y$ where the knowledge of $f_X$ is not needed to estimate $X$. This \emph{functional} dependence of $\widehat{x}$ on $f_Y$ was achieved for the Poisson, geometric, and Laplacian kernels  in~\citep{robbins1956empirical}, and it was later extended to the gaussian kernel by~\citep{miyasawa1961empirical}. 

This work builds on Miyasawa's result, and we repeat the calculation here in our notations. Take $X$ to be a random variable in $\mathbb{R}^d$ and $Y$ a noisy observation of $X$ with a known gaussian kernel with symmetric positive-definite $\Sigma \succ 0$:
\begin{equation}
\nonumber
	f(y|x) = \frac{1}{(2\pi)^{d/2} |\det(\Sigma)|^{1/2}} \exp \left(-(y-x)^\top \Sigma^{-1} (y-x)/2\right).
\end{equation}
It follows,
\begin{equation}
\nonumber
  \Sigma~\nabla_y f(y|x) = f(y|x) (x-y) .
\end{equation}
Multiply the expression above by $f(x)$ and integrate,
\begin{equation}\nonumber
  \Sigma~\nabla f(y) = \int (x-y) f(y|x) f(x) dx = f(y)(\widehat{x}(y)  - y),
\end{equation}
which then leads to the expression 
	$\widehat{x}(y) = y + \Sigma~\nabla \log f(y).$
For the isotropic case $\Sigma=\sigma^2 I_d$, the estimator takes the form
\begin{equation}
    \label{eq:NEBLS}
  \widehat{x}(y) = y + \sigma^2 \nabla \log f(y).
\end{equation}
To sum up, the estimator above is obtained in a setup where only the corrupted data (the random variable $Y$) are observed, with the knowledge of the measurement (gaussian) kernel \emph{alone} and without \emph{any knowledge} of the prior $f_X$. The remarkable result is the fact the least-squares estimator of $X$   is written in closed form purely as a functional of $\nabla \log f$, also known as the \emph{score function}~\citep{hyvarinen2005estimation}. The expression above for the \emph{least squares} estimator of $X$ is the basis for an algorithm to approximate $\nabla \log f$ which is discussed next.


In \emph{empirical Bayes}, the random variable $X$ is estimated in least squares from the noisy observations $Y$, without \emph{any} knowledge of the \emph{prior} $f_X$. But how did we get here? We started the paper with the i.i.d. sequence $X_1,\dots,X_n,$ where $Y$ did not even exist! \begin{quote} \it The idea is a  Gedankenexperiment of sort, where we first construct $Y= X+N(0,\sigma^2 I_d)$ by taking samples $Y_{ij} \sim f_X * f_N$; in turn, the experimenter (in the school of  empirical Bayes) ``observes'' $Y_{ij}$ and estimates $X$. But in this artificially supervised setup (the ``target'' is $X_i$), the error signal $\Vert X_i -\widehat{x}(Y_{ij})\Vert^2$ can also be measured\textemdash this is not the case in~\citep{robbins1956empirical}\textemdash and it drives the learning. The learning is achieved with stochastic gradient descent as the experimenter has parameterized the energy function with a neural network and can compute the stochastic gradients~\citep{robbins1951stochastic},
$$ \nabla_\theta \Vert X_i -Y_{ij}+ \nabla_y \phi(Y_{ij},\theta) \Vert^2. $$
Approximating $\nabla \log f$ in this manner is the essence of neural empirical Bayes. 
\end{quote}


For \emph{finite} samples, the best we can do is to {\it approximate} $\nabla \log f$. Given $X_1,\dots,X_n$, the i.i.d. samples $Y_{ij} \sim f_X*f_N$ are given by $ Y_{ij} = X_i + \varepsilon_j,\ \varepsilon_j \sim N(0,\sigma^2 I_d).$ These are indeed i.i.d. samples from density estimated by the kernel density estimator $$ \widehat{f}_Y(y) = \frac{1}{n} \sum_{i=1}^n f_N(y-X_i), $$
where the subscript in $f_N$ is a short for $N(0,\sigma^2 I_d)$. Here, $\sigma$ is a {\it hyperparameter} and the kernel estimator above is considered an estimator of $f_Y$. Note that the \emph{optimal bandwidth} in estimating $f_X$ depends on $n$~\citep{van2000asymptotic}.

The problem is that, in high dimensions, the kernel density estimator $\widehat{f}_Y$ (or any other {\it nonparametric estimator}) suffers from a severe {\it curse of dimensionality}~\citep{wainwright2019high}. This leads naturally to \emph{deep neural architectures} which has been viewed as an alternative for ``breaking the curse of dimensionality''~\citep{bengio2009learning}. In this work, the \emph{energy function} is parametrized with a deep neural network $\phi: \mathbb{R}^d \rightarrow \mathbb{R}$ with parameters $\theta$. The objective $\mathcal{L}(\theta)$ is then approximated, 
$$
        \mathcal{L}(\theta) \approx \sum \left\Vert X_i - Y_{ij}+\sigma^2 \nabla \energy(Y_{ij},\theta)\right\Vert^2,
$$
At  optimality (achieved with \emph{stochastic gradient descent}), $\theta^* = \argmin \mathcal{L}(\theta)$, 
$$ -\nabla \energy(\cdot,\theta^*) \approx \nabla \log f(\cdot).$$

\begin{remark}[implicit vs. explicit parameterization]  \label{remark:implicit} As it is made clear, our goal is to approximate $\nabla \log f: \mathbb{R}^d \rightarrow \mathbb{R}^d$. In doing so, another choice could have been the explicit parametrization of the score function as $\psi: \mathbb{R}^d \rightarrow \mathbb{R}^d$, where one avoids the computation $\nabla \phi$ in the optimization, but there, $\partial_j \psi_i = \partial_i \psi_j$ must hold pointwise. The parametrization of the energy function as $\phi:\mathbb{R}^d \rightarrow \mathbb{R}$ is essentially an implicit parametrization of $\nabla \log f$, since the computation $\nabla \phi$ is not a symbolic one, but achieved with automatic differentiation~\citep{baydin2018automatic}. Also,  the equivalence between the two strategies for one-hidden-layer networks~\citep{vincent2011connection} breaks down for two hidden layers and beyond. See~\citep{saremi2019approximating} for a proof and further discussions.	
\end{remark}

\begin{table}[h!]
\begin{center}
\begin{tabular}{ c|c|c|c|c} 
  & min &  median & mean & max~ $\approx \sigma_c$ \\ 
 \hline
 $\chi$ & $0.0186$ &  $0.1822$ & $0.1812$ & $0.2884$ \\ 
\end{tabular}
\end{center}
\caption{Some statistics of the matrix  $\chi$ (see Equation~\ref{eq:chi}) estimated from  $10^7$ pairs from the \emph{handwritten digit database}~\citep{lecun1998gradient}. The $\sigma$ in each experiment must be viewed in relation to these numbers and our review of the \emph{concentration of measure} in Section~\ref{sec:hdp} (see Figure~\ref{fig:twospheres}).}
\label{table:chi}
\end{table}

 \setcounter{subfigure}{0}
 \begin{figure}[h!]
 \vspace{0.5cm}
 \begin{center}
     \begin{subfigure}{\includegraphics[width=0.6\textwidth]{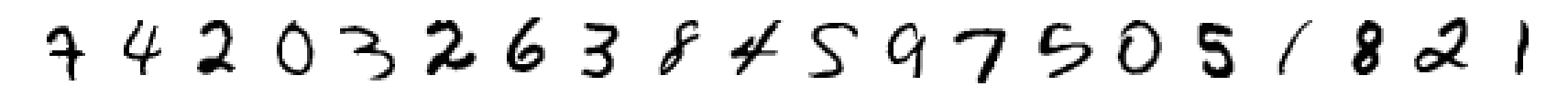}}%
     \vspace{-0.5cm}
     \end{subfigure}
     \vspace{-0.5cm}
     \begin{subfigure}{\includegraphics[width=0.6\textwidth]{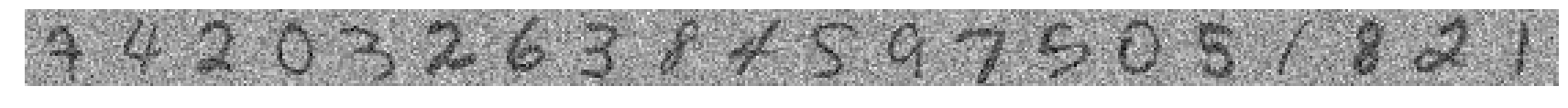}}%
     \end{subfigure}
     \begin{subfigure}{\includegraphics[width=0.6\textwidth]{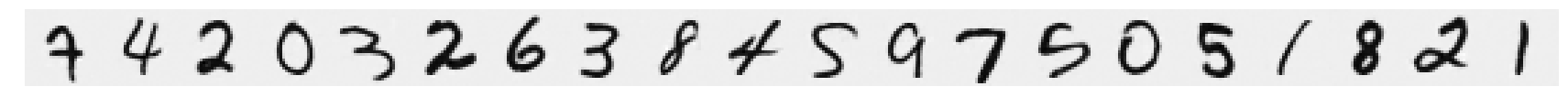}}%
     \end{subfigure} 
     \caption{ ($\sigma \approx \sigma_c$) Denoising performance of DEEN with a single jump for $\sigma=0.3$. The noisy pixel values are in the range  $\tt [-1.200, 1.995]$, and the denoised ones are in $\tt [-0.0749,1.0539]$. }
     \label{fig:denoising3e-1}   
 \end{center}
 \end{figure} 
  \setcounter{subfigure}{0}
 \begin{figure}[b!]
 \begin{center}
     \vspace{-0.5cm}
     \begin{subfigure}{\includegraphics[width=0.97\textwidth]{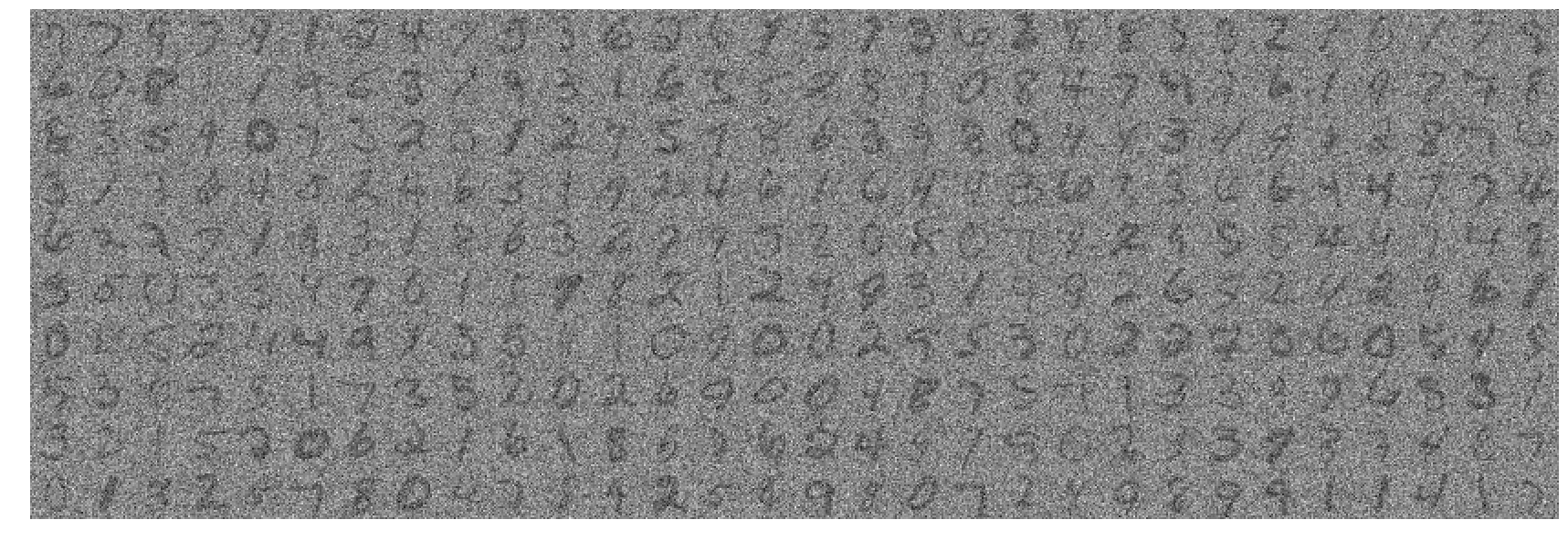}}%
     \end{subfigure}
     \begin{subfigure}{\includegraphics[width=0.97\textwidth]{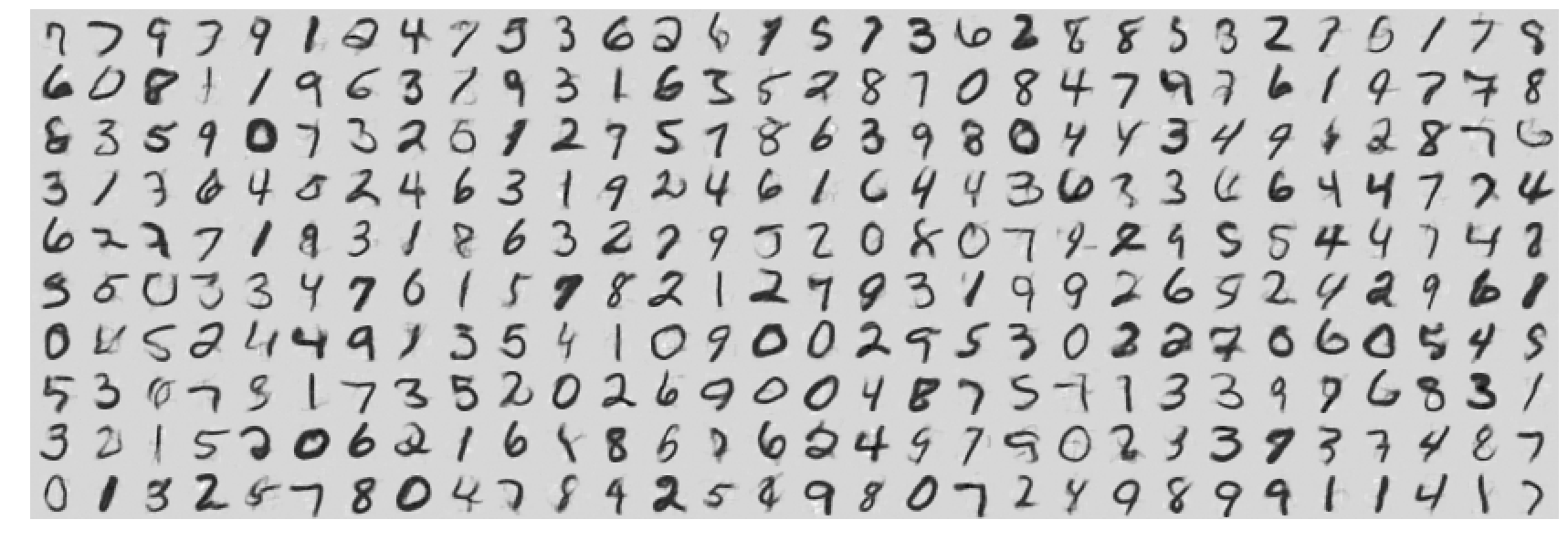}}%
     \end{subfigure}
     \caption{ ($\sigma > 2 \sigma_c$) Here, $\sigma=0.7$. The noisy pixel values are in the range $\tt [-3.314 , 3.683]$, and the denoised ones  are in $\tt [-0.2405, 1.2686]$. In this regime, the whole database is inside each $i$-sphere (see Figure~\ref{fig:twospheres}c).}
     \label{fig:denoising7e-1}   
 \end{center}
 \end{figure} 
\section{Extreme noise/denoising}
\label{sec:deendenoise}
In empirical Bayes, the least-squares estimator of $X$,
$$
\widehat{x}(y) = y + \sigma^2 \nabla \log f(y), $$
is derived for  any $\sigma$, which inspires us to call this a ``jump'' to capture the fact $\sigma$ may in fact be ``large''. Denoising is not our actual goal, but the expression above can be viewed as ``denoising $y$ to $\widehat{x}(y)$''. The denoising performance is important since  in the machinery of neural empirical Bayes, the goal of approximating the score function is formulated by the ``denoising objective'' $$\mathcal{L}(\theta)=\expectation \Vert X-Y+\sigma^2 \nabla \phi(Y,\theta)\Vert^2.$$As the least-squares estimator of $X$ is derived for any $\sigma$, the denoising performance of DEEN may also be tested to its limits. Here, we report such experiments for MNIST, where $d=784$ and $X$ is in the hypercube $[0,1]^d$ (see Remark~\ref{remark:gray}). Next, we elaborate on the geometric meaning of the noise levels, where in addition we  define a notion of ``extreme noise''.
\begin{itemize}
	\item $(\sigma>\sigma_c)$. The value $\sigma_c = \max_{ii'} \chi_{ii'}$ is defined as the onset of \emph{extreme noise}. Called ``extreme'', because for $\sigma>\sigma_c$, all $i$-spheres in the dataset overlap (see Figure~\ref{fig:twospheres}b). For MNIST, $\sigma_c \approx 0.2884 $ was estimated from $10^7$ handwritten digit pairs (see Table~\ref{table:chi}). The results for $\sigma=0.3$  on the \emph{test set} is presented in Figure~\ref{fig:denoising3e-1}. It should be stated that, this value of ``extreme noise'', just above $\sigma_c$, is not visually \emph{perceived} as extreme.
	\item $(\sigma>2\sigma_c)$. In this regime, due to geometry, the whole dataset is in \emph{inside} of each $i$-sphere (see Figure~\ref{fig:twospheres}c). This is very extreme! For MNIST, we experimented with $\sigma=0.7$ which is in this regime. The results on the \emph{test set} are presented in Figure~\ref{fig:denoising7e-1}.  
\end{itemize}

\begin{remark} \label{remark:gray}
  Note that the least-squares estimator of $X$ can take values anywhere in $\mathbb{R}^d$, not restricted to be in $[0,1]^d$. This is the reason for the ``gray background'' in the experiments, in relation to which, in the figure captions, we report the min/max of all the pixel values. This also holds for the jump samples presented in the next section.
\end{remark}

\subsection{The network architecture}
The following comments are in order regarding the \emph{network architecture} we used in our experiments.
\begin{itemize}
	\item (\emph{activation function}) The denoising results are a significant improvement over~\citep{saremi2018deep}. This was due to the use of ConvNets, instead of a fully connected network (more on that below) and the use of a \emph{``smooth ReLU''} activation function$$\sigma(z,\beta) = z/(1+\exp(-\beta z)),$$
 where the default was $\beta=1$. The activation function above converges uniformly to ReLU in the limit $\beta \rightarrow \infty$. It has been studied independently from different angles~\citep{elfwing2018sigmoid,ramachandran2017swish}. ReLU consistently gave a higher loss, which we believe is due to the term $\nabla_y \energy$ in the learning objective, which must be computed first before computing $\nabla_\theta \mathcal{L}$ to update the parameters $\theta \leftarrow \theta -\epsilon \nabla_\theta \mathcal{L}.$
	\item {\it (wide convnets)} All experiments reported in the paper were performed in a \emph{fixed} wide ConvNet architecture with the expanding $\rm channels=(256,512,1024)$, \emph{without pooling}, and with a \emph{bottleneck layer} of size $10$. All hidden layers were activated with the activation function $\sigma(\cdot,\beta=1)$, and the readout layer was linear.
	\item {\it (readout overparametrization)} We observed a slightly  faster convergence to lower loss (especially, very early in the training) by \emph{overparameterizing} the linear output layer. These experiments were inspired by the ``acceleration effects'' studied for \emph{linear neural networks}~\citep{arora2018optimization}, but we did not study it thoroughly. 
	\item  We used the \emph{Adam optimizer}~\citep{kingma2014adam}. The optimization was stable over a wide range of \emph{mini-batch sizes} and \emph{learning rates}, and as expected, we did not observe the validation/test loss going up in the experiments. This stability is due to the fact that the inputs to $\phi$ are noisy samples.  The \emph{automatic differentiation}~\citep{baydin2018automatic} was implemented in PyTorch~\citep{paszke2017automatic}. DEEN is ``memory hungry'' but the computational costs are not addressed here in detail. 
\end{itemize}

\section{Walk-jump sampling: Langevin walk,  Robbins jump}
\label{sec:walkjump}
Sampling ``complex distributions'' in high dimensions is an intractable problem due to the fact that their probability mass is concentrated in \emph{sharp ridges} that are separated by \emph{large regions of low probability}, therefore making MCMC methods ineffective \emph{(the ``low-dimensional manifold'' $\mathscr{M}$ is in fact quite complex when viewed in the ambient space $\mathbb{R}^d$)}. These problems are well known, but empirically they have mostly been studied for sampling $f_X$. Regarding the random variable $Y$, it is clear that the problem of sampling $f=f_X*f_N$ is in fact easier. The idea is that MCMC mixes faster as $\mathscr{N}$ is less complex. The ``sharp ridges'' are smoothed out and (as we argued in Section~\ref{sec:hdp}) $\mathscr{M}$ itself is expanded in dimensions, $\text{dim}(\mathscr{N})\gg \text{dim}(\mathscr{M})$, therefore the ``large regions of low probability'' themselves are smaller in size. In summary, \emph{Langevin MCMC} is believed to be more effective in sampling $$\exp(-\phi)/Z\approx f$$
than $f_X$ (putting aside the harder problem of learning $\nabla\log f_X$). At any time, the estimator $\widehat{x}$ can be used to \emph{jump} near $\mathscr{M}$. ``Near'' is intuitive, and it should be stated that neither $f_{X|Y}$ nor $f_X$ is exactly sampled during jumps; the jump samples must be seen as heuristic. 

In what follows, it is understood that $\phi$ is at optimality. We first give a description of the algorithm and then express the equations.\begin{itemize}
\item {\it(walks)}	The Langevin MCMC is used to draw statistical samples $y_t \sim \exp(-\phi)/Z.$  Langevin MCMC is based on discretizing the \emph{Langevin diffusion}~\citep{van1992stochastic} and  the updates are based solely on $\nabla \phi$ (the equations are coming next), therefore not knowing the \emph{normalizing constant} $Z$ is of no concern.
\item {\it(jumps)} Given $y_\tau$, ideally an \emph{exact sample}~\citep{mackay2003information} from $\exp(-\phi)/Z$, the jumps are made with the \emph{Bayes estimator} of $X$. The \emph{spirit of empirical Bayes} is fully present here: the  jump samples are generated without knowing $\nabla f_X$  (or its approximation), and without running Langevin MCMC on its ``rougher terrain''.
\end{itemize}


What emerges is a sampling algorithm in two parts. The  Langevin MCMC is the \emph{engine} that should run continously. By contrast, the jumps can be made at any time. That is:
\begin{itemize}
	\item[(i)]  Sample from $\exp(-\energy)/Z$ with Langevin MCMC,
\begin{equation}
\label{eq:walk}\nonumber
	y_{t+1} = y_t - \delta^2 \nabla \energy(y_t) + \sqrt{2}  \delta \varepsilon,~\varepsilon \sim N(0, I_d),
\end{equation}
where $\delta \ll 1$ is the step size and here $t$ is discrete time.
\item[(ii)]  At any time $\tau$, use the Bayes estimator of $X$ to jump from $y_\tau$ to $\widehat{x}(y_\tau)$,
\begin{equation}
	\label{eq:jump} \nonumber
	\widehat{x}(y_\tau) = y_\tau - \sigma^2 \nabla \energy(y_\tau).
\end{equation}
\end{itemize} 

Two types of approximations have been made in walk-jump sampling. The first is that the Langevin MCMC does not sample $f$ but the density $\exp(-\phi)/Z \approx f$; this approximation is unavoidable. The approximation involved in jumps are less understood. Stepping back, $\widehat{x}(y_\tau)$ {\it can} be thought of as approximating the {\it posterior} $f_{X|y_\tau}$ with a single Dirac mass at $\expectation(X|Y=y_\tau)$ but this is a very poor approximation to the posterior. Regarding $f_X$ itself, the intuitive idea behind Robbins jump is that the \emph{Bayes estimator} of $X$ ``lands'' near $\mathscr{M}$. But this intuition must be validated by  computing the covariance of the estimator, and this is indeed the next step to better understand the walk-jump sampling.

\begin{figure}[h!]
\centering
\includegraphics[width=0.93\textwidth]{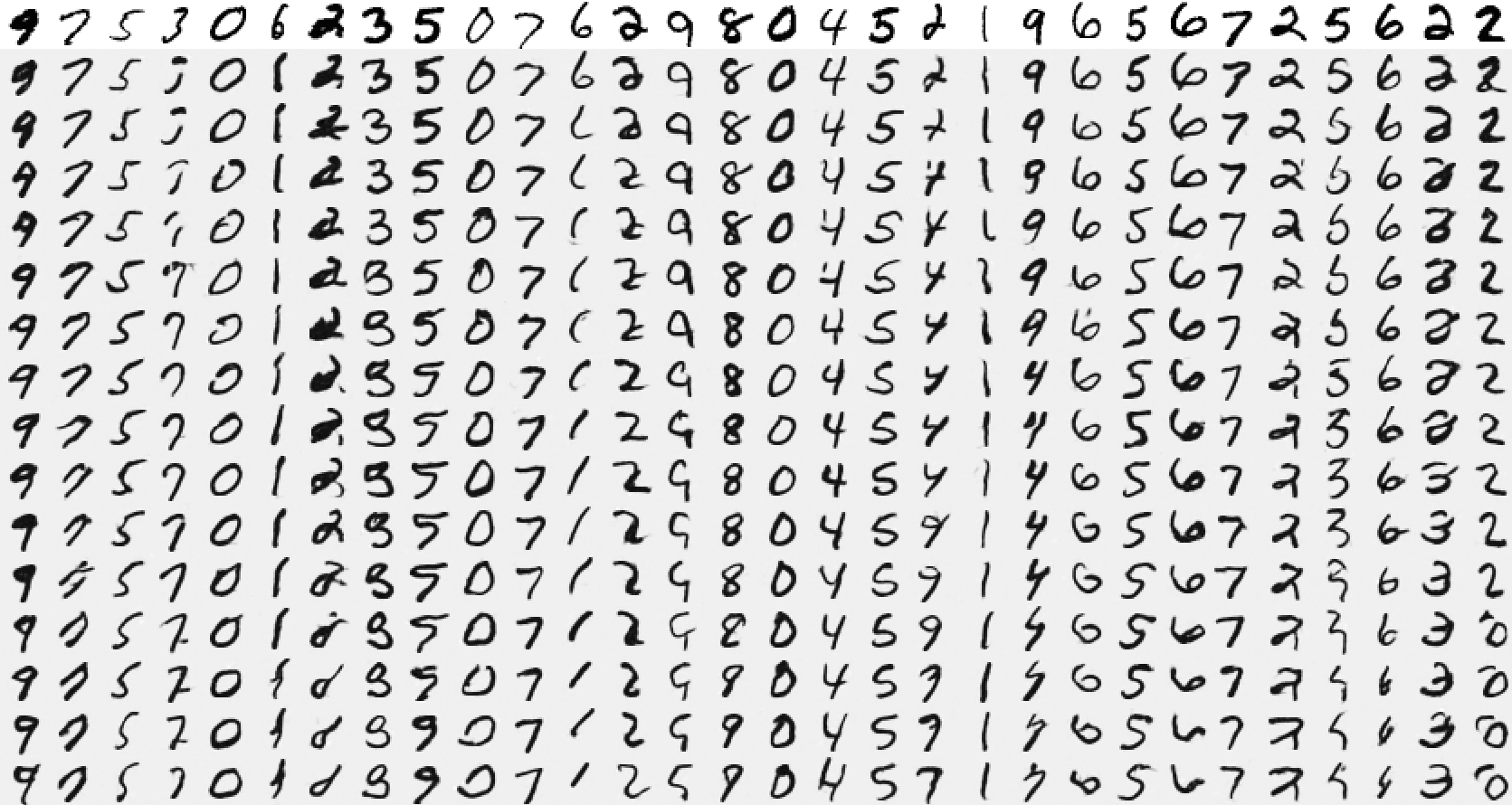}%
\caption{\emph{($\sigma\approx \sigma_c$)} Top row is $x_0\sim f_X$, sampled from the handwritten digit database which DEEN was trained on. The \emph{Langevin MCMC} was intialized at $y_0 = x_0 +\varepsilon$, $\varepsilon \sim N(0,\sigma^2 I_d)$, where $\sigma$ was the same value of $\sigma$ which DEEN had been trained on. The samples $y_t$ are are not shown. The \emph{jumps} are shown in multiples of $\tau_0 = 10^4$, and the step size was $\delta = \sigma/100.$ Here, $\sigma=0.3$. The pixel values are in the range  {\tt[-0.07, 1.10]}. }
\label{fig:run163mcmc}
\end{figure}
\begin{figure}[h!]
\centering
\vspace{0.5cm}
\includegraphics[width=0.93\textwidth]{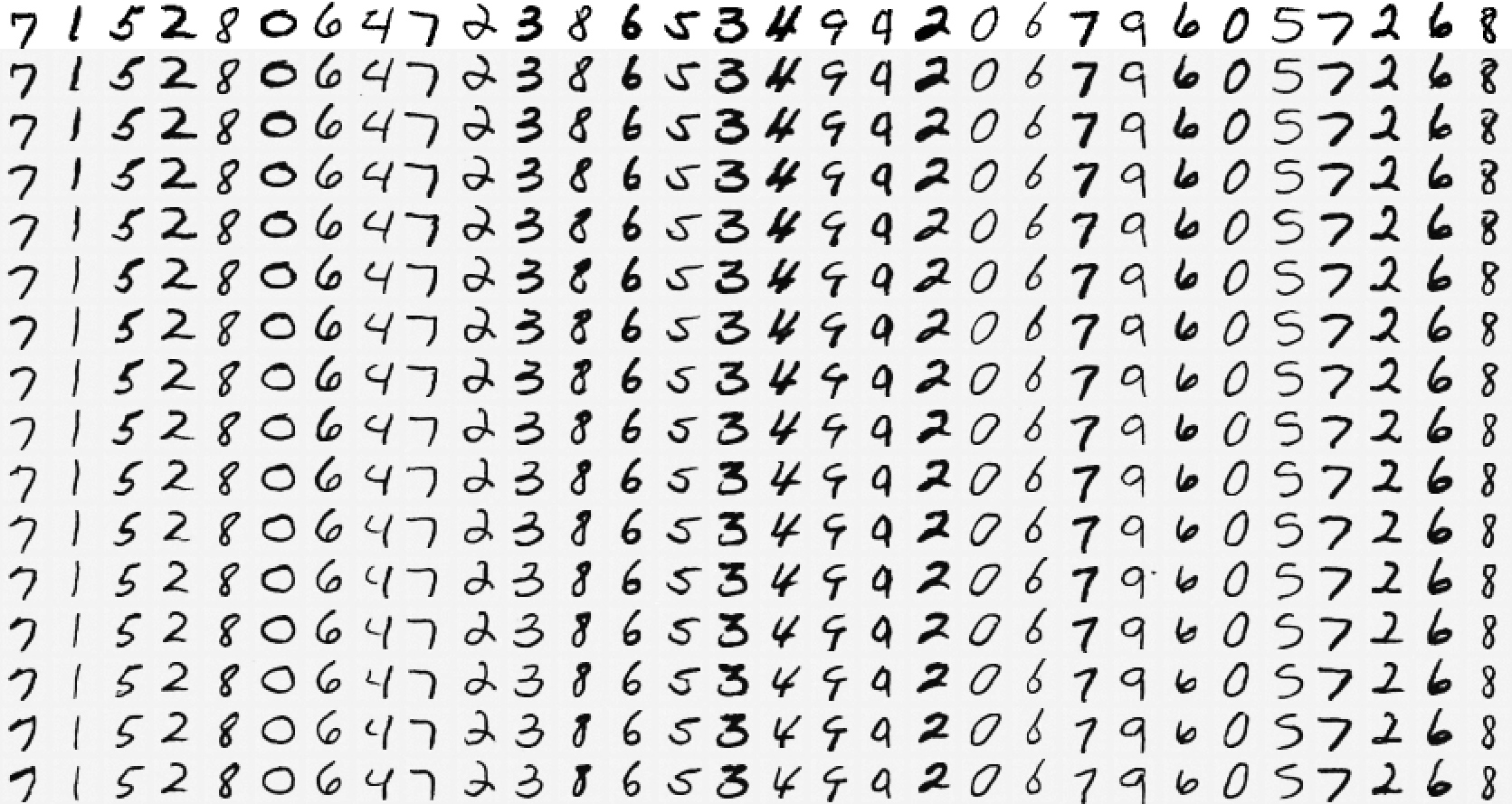}%
\caption{($\sigma< \text{median}(\chi)$) Here, $\sigma=0.15$, which is in the regime of mostly non-overlapping spheres. The Langevin MCMC parameters are set as above. The pixel values are in the range {\tt[-0.05, 1.04]}. }
\label{fig:run161mcmc}
\end{figure}

We tested the algorithm on the \emph{handwritten digit database} for $\sigma=0.3$ and $\sigma=0.15$. The results are shown in Figure~\ref{fig:run163mcmc} and~\ref{fig:run161mcmc} respectively. For $\sigma=0.3$ (in the regime of highly overlapping spheres), there was mixing between styles/classes. For $\sigma=0.15$ (in the regime of mostly non-overlapping spheres), samples stayed within a class while the styles changed. 

Some side-by-side comparisons with  {\it denoising autoencoders} are in order.
\begin{itemize}	
	\item In walk-jump sampling, the key step is sampling from the density $\exp(-\phi)/Z$. The least-squares estimation of $X$\textemdash the jumps\textemdash are decoupled from Langevin MCMC. This is in contrast to the chain $X_t \rightarrow Y_t \rightarrow X_{t+1}$ constructed in {\it generalized denoising autoencoders}~\citep{bengio2013generalized}, which does not suffer from the problems we mentioned for the jumps.
	\item Here, the learning objective is derived for \emph{any} $\sigma$.  By contrast, \emph{denoising autoencoders} approximate the score function  only in the limit $\sigma \rightarrow 0$~\citep{alain2014regularized}.  \emph{Generalized  denoising autoencoders}~\citep{bengio2013generalized} and {\it generative stochastic networks}~\citep{alain2016gsns} were devised as a remedy for those $\sigma \rightarrow 0$ limitations. This is avoided altogether here.
			\item In this work, there is a clear \emph{geometrical} notion for large and small $\sigma$, which is based on the concentration of measure phenomenon. This picture is lacking in the references cited, and it is not clear which ranges of noise values should be considered there. 
	\item Most importantly, denoising autoencoders suffer from \emph{``limited parameterization''} as expressed by~\citet{alain2014regularized}, and independently observed in~\citep{saremi2018deep}. To summarize, in denoising autoencoders, one must learn a curl-free encoder-decoder to be able to properly approximate the \emph{score function} and this is problematic beyond \emph{one-hidden-layer} architectures. In our work, this problem is avoided due to the \emph{energy parameterization} (see Remark~\ref{remark:implicit}).
\end{itemize}

\clearpage

\section{Neural empirical Bayes associative memory (NEBULA)} 
\label{sec:NEBULA}
In this section, the neural empirical Bayes machinery is used to define a new notion of associative memory. \emph{Associative memory} (also known as \emph{content-addressable memory}) is a deep subject in neural networks with a rich history and with roots in psychology and  neuroscience  (long-term potentiation and Hebb's rule). The depiction of this computation is even present in the arts and literature, championed by Marcel Proust, in the form of ``stream of consciousness'', and the constant back and forth between \emph{perception} and \emph{memory}. 

In 1982, Hopfield brought together earlier attempts to formulate \emph{associative memory}, and showed that the collective computation of a system of neurons with symmetric weights, in the form of asynchronous updates of \emph{McCulloch-Pitts neurons}~\citep{mcculloch1943logical}, minimize an energy function~\citep{hopfield1982neural}. The energy function was constructed in a Hebbian fashion. The associative memory was then formulated as the ``flow'' (the neurons were binary) to the local minima of the  energy function. 

However, Hopfield's energy function is not {\it the} energy function\footnote{In addition, Hopfield networks have severe limitations in \emph{memory capacity}, addressed most recently in~\citep{hillar2014robust, krotov2016dense,chaudhuri2017associative}.}: it does not approximate (learn) the negative log probability density function of its stored memories. The \emph{Boltzmann machine}~\citep{hinton1986learning} was developed in fact inspired by that observation, in which learning \emph{the} energy function was achieved by introducing {\it hidden units}. Regarding the associative memory and the \emph{phase space flow}, the problem is that in Boltzmann machines, hidden units need to be inferred first before having any notion of flow for the \emph{visible units}. And inference is indeed computationally very expensive\textemdash \emph{``the curse of inference''} (but not nearly as fundamental as the \emph{curse of dimensionality})\textemdash in \emph{probabilistic graphical models}~\citep{wainwright2008graphical,koller2009probabilistic}.

What we have achieved so far is to learn a function $\phi$ which approximates  \emph{the energy function} of $Y$, $\phi \approx -\log f$ (modulo a constant). The key here is that $\phi$ is a \emph{function} (a ``computer'') that {\it computes} $f(y)$ for any $y$; the ``hidden units'' in $\phi$ are not inferred. In other words, the hidden units (and the parameters) in $\phi$ are there solely for {\it universal approximation}, $-\nabla \phi \approx \nabla \log f.$  
In \emph{Boltzmann machines}, the hidden units are there to think! And this is the big difference. The definition below should be viewed against this backdrop.
 


\begin{definition}
We define the neural empirical Bayes associative memory (NEBULA), \`{a} la ~\citet{hopfield1982neural}, as the flow to strict local minima of the energy function $\energy$. In continuous time, the memory dynamics is governed by the gradient flow: \< y'(t) = -\nabla \energy(y(t)). \> 
NEBULA is identified by its attractors, the set of all the strict local minima of $\energy$:
$$ \mathcal{X}^* = \{ X_1^*,\dots,X_A^*\}.$$
The mapping $\mathscr{A}: \mathbb{R}^d \rightarrow \mathcal{X}^*$ denotes the convergence to an attractor under the gradient flow. The basin of attraction for the attractor $X_a^*$ is denoted by $\mathcal{A}(X_a^*)$ and defined (intuitively) as the largest subset of $\mathbb{R}^d$ such that (in a slight abuse of notation) $\mathscr{A}(\mathcal{A}(X_a^*))=\{X_a^*\}$.
\end{definition}

\begin{figure}[h!]
\centering
\includegraphics[width=\textwidth]{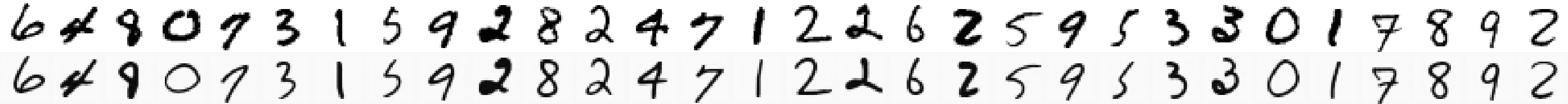}%
\caption{ {\it ($\sigma$=0.15)} The top row are $X_i$ from the MNIST test set. The bottom row are the attractors $\mathscr{A}(X_i)$; \emph{they are not statistical samples from $f_X$: $\mathscr{A}(X_i) \not\sim f_X$. }}
\label{fig:ram1}
\end{figure}
\begin{figure}[h!]
 \begin{center}
    \begin{subfigure}{\includegraphics[width=0.88\textwidth]{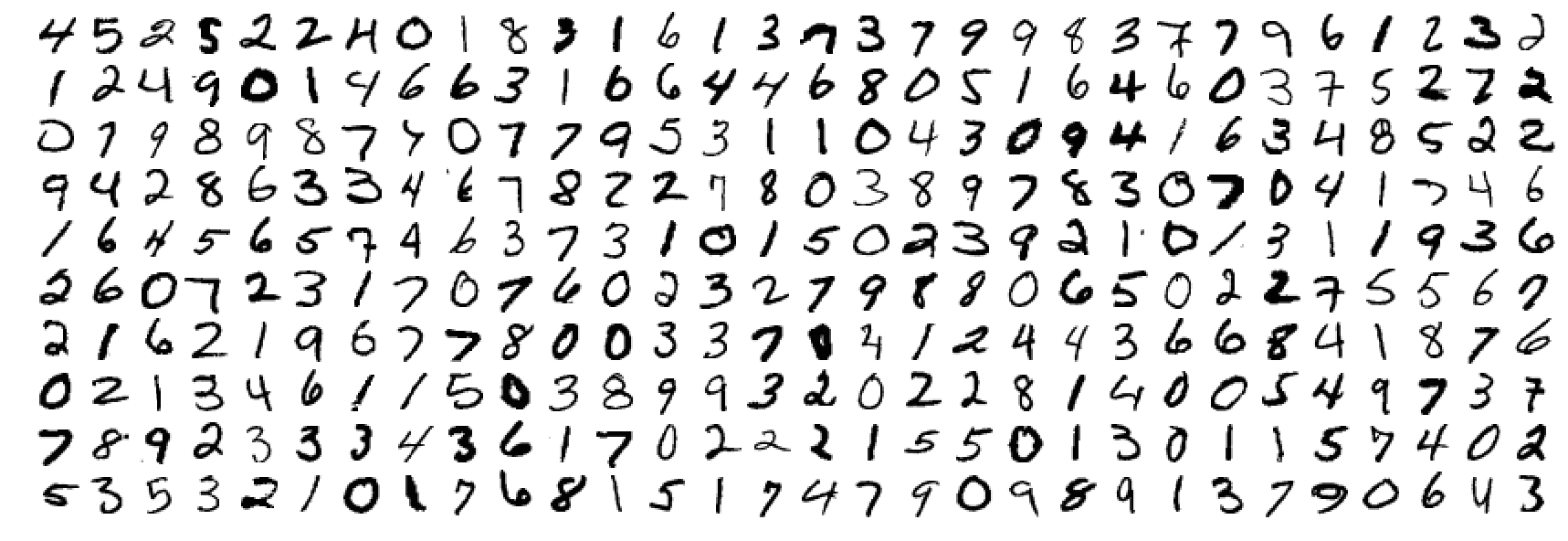}}%
    \vspace{-0.3cm}
    \end{subfigure}

    \begin{subfigure}{\includegraphics[width=0.88\textwidth]{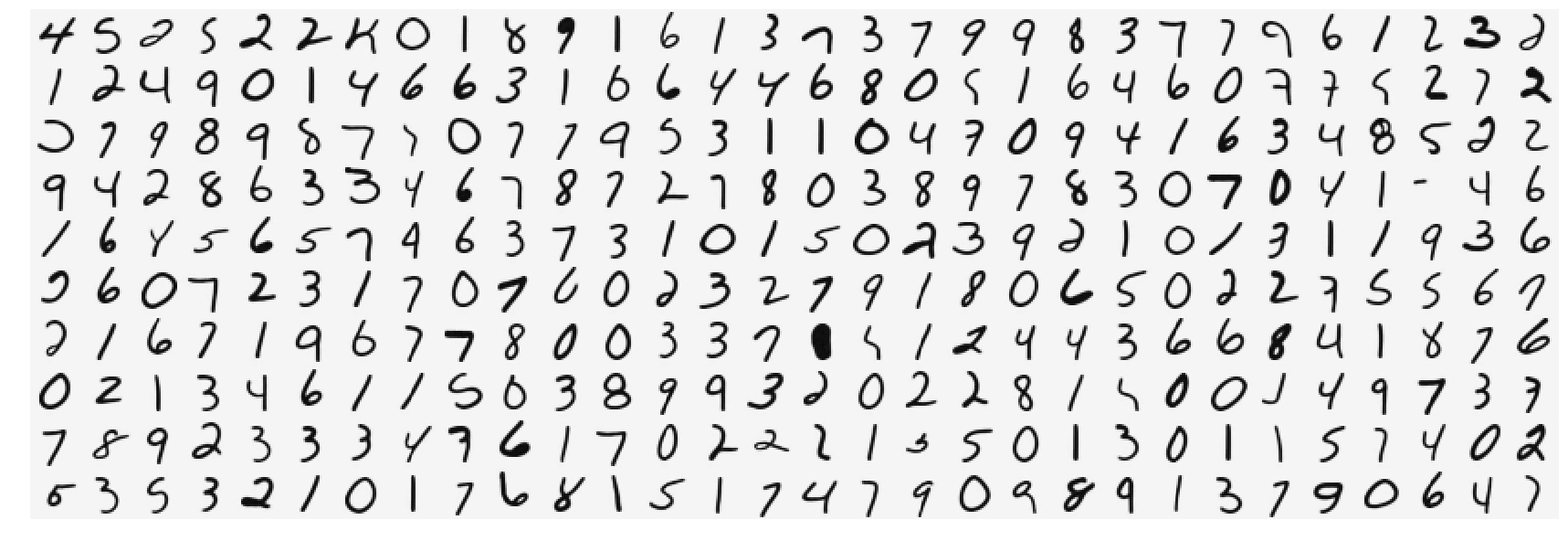}}%
    \end{subfigure}
    \caption{ {\it ($\sigma$=0.3)} As expected, $i$-sphere interactions are stronger compared to $\sigma=0.15$. Note the bottom right digit, flowing from a ``3'' to a ``7''.}  
    \label{fig:ram2}    
 \end{center}
 \end{figure}

What is so special about the local minima of $\energy$? Start with a \emph{single sample} $X_1$, $$\energy(y) = \Vert y-X_1 \Vert^2/(2\sigma^2),~ \mathcal{X}^*=\{X^*_1\},~\ X_1^*=X_1,~ \mathscr{A}(\cdot)=X_1,~ \mathcal{A}(X_1) =\mathbb{R}^d.$$
For \emph{many samples},  the learning objective is \emph{optimized} such that $-\nabla \energy$ evaluated on  $i$-spheres point to $X_i$ (see Figure~\ref{fig:gaussians}c), but there are \emph{conflicts of interests}, and these ``conflicts'' are the essence of $i$-sphere interactions, stated in Definition~\ref{def:isphere}; in its practical summary, larger $\sigma$ means more $i$-sphere overlaps, and therefore larger ``interaction couplings''. \emph{For NEBULA, the presence of $i$-sphere interactions implies that $X_a^*$ are not statistical samples from $f_X$.} In other words, given $X_i \sim f_X$, $\mathscr{A}(X_i) \not\sim f_X$. But what is the ``right metric'' $ d(X_i,\mathscr{A}(X_i)),$
to measure the distance between $X_i \sim f_X$ and $\mathscr{A}(X_i)$? The problem is that the flow is the gradient flow of $\phi\approx -\log f$ but the attractors are roughly speaking ``close to $\mathscr{M}$''. The total length of the path taken by the gradient flow is a natural choice but we leave that for furture studies. Our expectation (based on $i$-sphere interactions) is that $d(X_i,\mathscr{A}(X_i))$ will be larger for larger $\sigma$, visualized in Figures~\ref{fig:ram1} and~\ref{fig:ram2}, with pronounced qualitative differences. 

 \setcounter{subfigure}{0}



\begin{figure}[h!]
\centering
\includegraphics[width=0.97\textwidth]{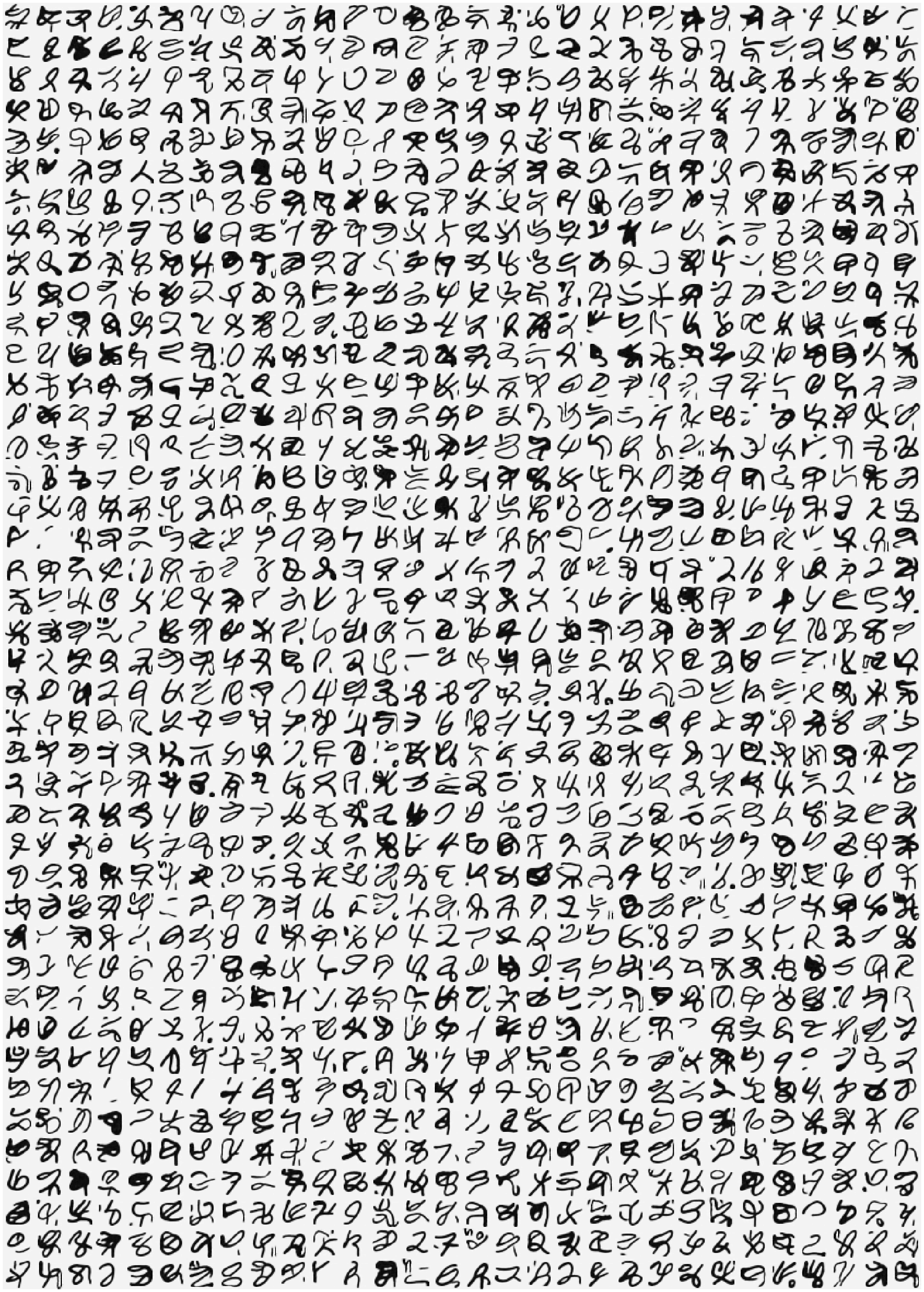}%
\caption{Attractors of NEBULA for $\sigma=0.3$ ($\sigma_c \approx 0.29$).}
\label{fig:nebula1}
\end{figure}	

\begin{figure}[h!]
\centering
\includegraphics[width=0.97\textwidth]{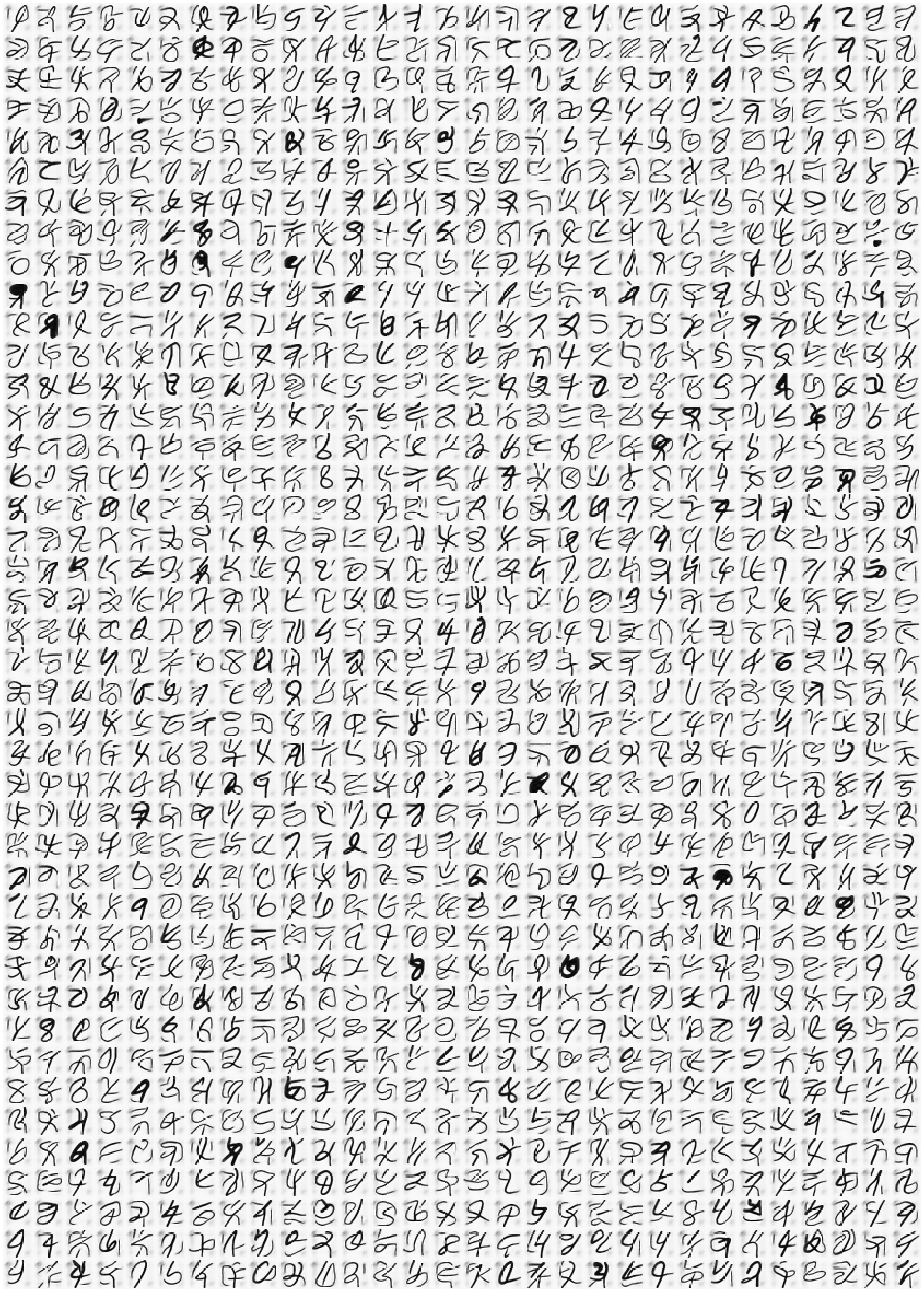}%
\caption{Attractors of NEBULA for $\sigma=0.25$ ($\sigma_c \approx 0.29$).}
\label{fig:nebula2}
\end{figure}	

\clearpage

\section{Emergence of ``creative memories''}
\label{sec:creative}
We finish the paper with some surprising results on the emergence of highly-structured non-digit memories, named \emph{``creative memories''}. The experiments designed in this section are a natural continuation of the experiments from the previous section. In its summary, NEBULA is ``well behaved'' when it is initialized at $X_i \sim f_X$. \emph{But what if it is not initialized as such?} Here, we explore \emph{$i$-sphere interactions} (see Definition~\ref{def:isi}) from a different angle by initializing NEBULA at a random point, $$ y_0 \sim \text{Unif}([0,1]^d).$$ The results in Figures~\ref{fig:nebula1} and~\ref{fig:nebula2} are the strongest evidence for \emph{$i$-sphere interactions} as being a good \emph{abstraction}. In viewing Figures~\ref{fig:nebula1} and~\ref{fig:nebula2}, consider Figure~\ref{fig:twospheres}, but with \emph{60000} highly overlapping $i$-spheres, and $-\nabla \phi$ on each $i$-sphere that should ``point towards'' $X_i$ \emph{in expectation}, where in addition, there are other complexities (touched upon in Remark~\ref{remark:uniform}) regarding the $i$-spheres themselves.

Several remarks are in order.
\begin{itemize}
	\item The images shown in Figures~\ref{fig:nebula1} and~\ref{fig:nebula2} are \emph{not} statistical samples from $f_X$ nor $f_Y$.
	\item They are not the result of mass aggregation of the \emph{kernel density estimator} $\widehat{f}_Y$. {\it Mode seeking} is a deep topic in the kernel density literature around the {\it mean shift algorithm}~\citep{fukunaga1975estimation, cheng1995mean,comaniciu2002mean}, also around the intriguing topic of ``ghost modes''~\citep{carreira2003number}. Of course, the \emph{attractors} reported here has not been reported in the kernel density literature (see~\citep{carreira2015review} and the examples therein for MNIST).
	\item The results presented are from all the random initializations in the \emph{unit hypercube} from which we ran the algorithm, without any hand-picking!
\end{itemize}




%

\section{Summary} 
\label{sec:summary}
In \emph{neural empirical Bayes}, the smoothing of a random variable $X$ to $Y$ ($f=f_X*f_N$), denoising, and the (unnormalized) density estimation of the smoothed density $f$ were unified in a single machinery. Its inner workings were captured symbolically by $X \rightleftharpoons Y$, as well as by a \emph{Gedankenexperiment} with an ``experimenter'' in the school of~\citet{robbins1956empirical} and~\citet{robbins1951stochastic}.\footnote{In our presentation, the smoothing part was \emph{viewed} from the angle of \emph{kernel density estimation}, with an important distinction that the kernel bandwidth was a \emph{hyperparameter}. But on reflection, 
this is not necessary. Fundamentally, setting aside the ``\emph{neural} energy function'', one can go on with our program by just knowing the (deep) machineries of  {\it empirical Bayes}~\citep{robbins1956empirical} and \emph{stochastic gradients}~\citep{robbins1951stochastic}.} In this machinery, the energy function is parametrized with a neural network and SGD becomes the engine for learning, whose end result is captured by $\nabla \phi(\cdot,\theta^*) \approx -\nabla \log f(\cdot)$. We proposed an approximative walk-jump sampling scheme which produces samples which are particularly appealing visually due to the denoising jump used. The energy function can further be used as an associative memory (NEBULA), which even has some ``creative'' capacities.

 


\section*{Acknowledgement} 
This work was supported by the Canadian Institute for Advanced Research, the Gatsby Charitable Foundation, and the National Science Foundation through grant IIS-1718991. We especially thank Bruno Olshausen, Eero Simoncelli, and Francis Bach for discussions.

\bibliography{19-216.bib}

\end{document}